\documentclass[10pt,twocolumn,letterpaper]{article}

\usepackage{3dv}
\usepackage{times}
\usepackage{epsfig}
\usepackage{graphicx}
\usepackage{amsmath}
\usepackage{amssymb}

\usepackage{xcolor}
\usepackage{tabularx}
\usepackage{rotating}
\usepackage{verbatim}
\usepackage{xcolor,colortbl}
\usepackage{caption}
\usepackage{etoolbox}
\usepackage[normalem]{ulem}
\usepackage{booktabs}
\usepackage{caption}
\usepackage{subfig}

\definecolor{car}{rgb}{0.39215686, 0.58823529, 0.96078431}
\definecolor{bicycle}{rgb}{0.39215686, 0.90196078, 0.96078431}
\definecolor{motorcycle}{rgb}{0.11764706, 0.23529412, 0.58823529}
\definecolor{truck}{rgb}{0.31372549, 0.11764706, 0.70588235}
\definecolor{other-vehicle}{rgb}{0.39215686, 0.31372549, 0.98039216}
\definecolor{person}{rgb}{1.        , 0.11764706, 0.11764706}
\definecolor{bicyclist}{rgb}{1.        , 0.15686275, 0.78431373}
\definecolor{motorcyclist}{rgb}{0.58823529, 0.11764706, 0.35294118}
\definecolor{road}{rgb}{1.        , 0.        , 1.        }
\definecolor{parking}{rgb}{1.        , 0.58823529, 1.        }
\definecolor{sidewalk}{rgb}{0.29411765, 0.        , 0.29411765}
\definecolor{other-ground}{rgb}{0.68627451, 0.        , 0.29411765}
\definecolor{building}{rgb}{1.        , 0.78431373, 0.        }
\definecolor{fence}{rgb}{1.        , 0.47058824, 0.19607843}
\definecolor{vegetation}{rgb}{0.        , 0.68627451, 0.        }
\definecolor{trunk}{rgb}{0.52941176, 0.23529412, 0.        }
\definecolor{terrain}{rgb}{0.58823529, 0.94117647, 0.31372549}
\definecolor{pole}{rgb}{1.        , 0.94117647, 0.58823529}
\definecolor{traffic-sign}{rgb}{1.        , 0.        , 0.    }    

\makeatletter
\newcommand{\car@semkitfreq}{3.92}
\newcommand{\bicycle@semkitfreq}{0.03}
\newcommand{\motorcycle@semkitfreq}{0.03}
\newcommand{\truck@semkitfreq}{0.16}
\newcommand{\othervehicle@semkitfreq}{0.20}
\newcommand{\person@semkitfreq}{0.07}
\newcommand{\bicyclist@semkitfreq}{0.07}
\newcommand{\motorcyclist@semkitfreq}{0.05}
\newcommand{\road@semkitfreq}{15.30}  %
\newcommand{\parking@semkitfreq}{1.12}
\newcommand{\sidewalk@semkitfreq}{11.13}  %
\newcommand{\otherground@semkitfreq}{0.56}
\newcommand{\building@semkitfreq}{14.1}  %
\newcommand{\fence@semkitfreq}{3.90}
\newcommand{\vegetation@semkitfreq}{39.3}  %
\newcommand{\trunk@semkitfreq}{0.51}
\newcommand{\terrain@semkitfreq}{9.17} %
\newcommand{\pole@semkitfreq}{0.29}
\newcommand{\trafficsign@semkitfreq}{0.08}
\newcommand{\semkitfreq}[1]{{\csname #1@semkitfreq\endcsname}}

\usepackage[pagebackref=true,breaklinks=true,letterpaper=true,colorlinks,bookmarks=false]{hyperref}

\threedvfinalcopy %

\def\threedvPaperID{140} %

\ifthreedvfinal\fi
\begin{document}

\title{LMSCNet: Lightweight Multiscale 3D Semantic Completion}

\author{Luis Rold\~ao\\
Inria, 
AKKA Technologies\\
{\tt\small luis.roldao@inria.fr}
\and
Raoul de Charette\\
Inria\\
{\tt\small raoul.de-charette@inria.fr}
\and
Anne Verroust-Blondet\\
Inria\\
{\tt\small anne.verroust@inria.fr}
}

\twocolumn[{%
\renewcommand\twocolumn[1][]{#1}%
\maketitle

\begin{center}
	\centering
	\includegraphics[width=1\textwidth]{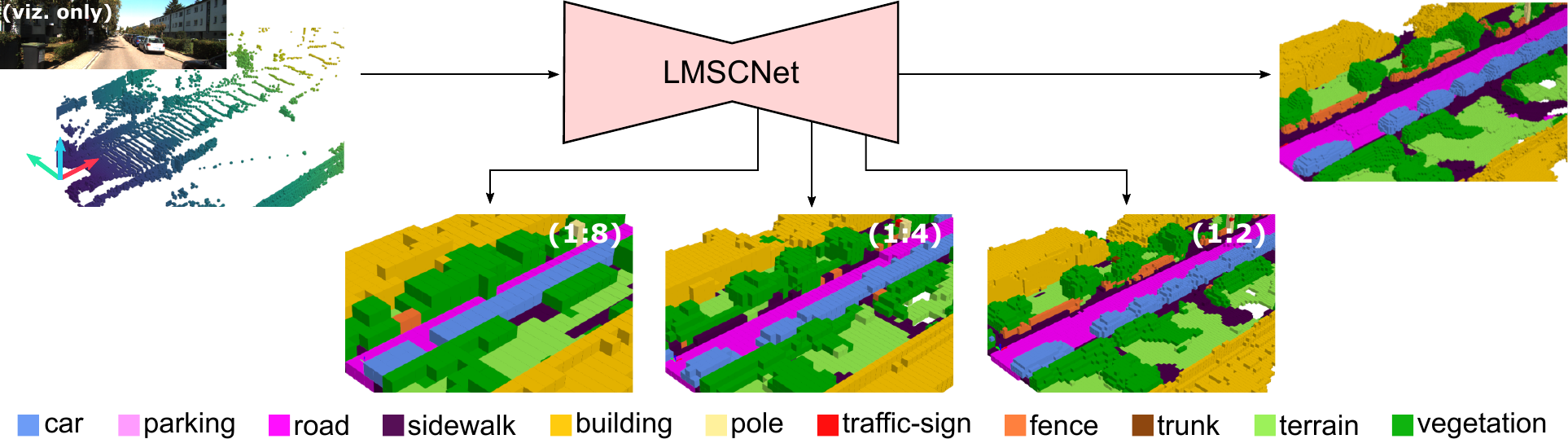}
	\captionof{figure}{To prevent heavy computation overhead we use a mix of 2D/3D convolutions to infer multiscale 3D semantic scene completion from sparse voxelized input. Evaluation performed on the challenging SemanticKITTI~\cite{Behley2019SemanticKITTIAD} benchmark shows that our LMSCNet proposal reaches state-of-the-art performance at significantly faster computation speed.}
	\label{fig:intro}
\end{center}
}]

\maketitle

\begin{abstract}
We introduce a new approach for multiscale 3Dsemantic scene completion from voxelized sparse 3D LiDAR scans.
As opposed to the literature, we use a 2D UNet backbone with comprehensive multiscale skip connections to enhance feature flow, along with 3D segmentation heads. 
On the SemanticKITTI benchmark, our method performs on par on semantic completion and better on occupancy completion than all other published methods -- while being significantly lighter and faster. As such it provides a great performance/speed trade-off for mobile-robotics applications. 
The ablation studies demonstrate our method is robust to lower density inputs, and that it enables very high speed semantic completion at the coarsest level. 
Our code is available at \url{https://github.com/cv-rits/LMSCNet}.
\end{abstract}

\section{Introduction}
\label{sec:intro}

Understanding 3D surroundings is a natural ability for humans. While past experience allows us to reason about scene geometry and semantics of an entire scene, this proves difficult for computers given the inherently sparse nature of 3D sensors~\cite{Berger2017ASO} (due to sparse sensing, limited field-of-view, and occlusions). 
Still, a comprehensive 3D sensing of the scene is crucial for applications like mobile robotics, and more especially for autonomous driving.
Recently, semantic scene completion was proposed~\cite{Song2017SemanticSC} as a new generative task, where both completion and semantic labels are inferred for the whole scene.

Unlike images, conveniently encoded as 2D tensors, 3D data causes representation challenges.
It is thus common to encode the latter as voxel grids processed by 3D Convolutional Neural Networks (CNNs)~\cite{Dai2017ShapeCU, Song2017SemanticSC, Dai2018ScanCompleteLS, Liu2018SeeAT}. 
This shows good results but also requires heavy computation, as the memory requirement grows cubically with the input voxel resolution~\cite{Liu2019PointVoxelCF}. 
Consequently, most of the literature limits the predicted resolution and network depth, being incapable to perform the task at the same spatial resolution as the input~\cite{Song2017SemanticSC, Garbade2019TwoS3, Zhang2018EfficientSS}. 
This drawback has limited the deployment of such methods for real time applications -- $i.e.$ augmented and virtual reality~\cite{Krevelen2010ASO}, robotics perception and navigation~\cite{Kober2013ReinforcementLI}, scene understanding~\cite{Li2009TowardsTS}, among others -- that would greatly benefit of semantic scene completion from sparse LiDAR scan.

We tackle this problem, and propose a Lightweight Multiscale Semantic Completion, coined LMSCNet, where a 3D voxel grid is processed with considerably lighter 2D convolutions without need of additional modalities~\cite{Behley2019SemanticKITTIAD,Garbade2019TwoS3}.
This is achieved by convolving along one spatial axis (close in spirit to bird-eye view process~\cite{Chen2018DeepLabSI}), while mapping to third dimension with 3D segmentation heads. 
In our proposal, multiscale completion is also possible given informative features map flow, preserving computation efficiency and enabling very fast inference at coarse levels.
Fig.~\ref{fig:intro} shows the multiscale output of our LMSCNet on the SemanticKITTI dataset~\cite{Behley2019SemanticKITTIAD}, using a single sparse LiDAR scan input encoded as voxel.
While some works use progressive multiscale losses~\cite{Li2009TowardsTS,Dai2019SGNNSG, Dai2018ScanCompleteLS}, the literature ignores the benefit of multiscale completion which we prove useful for reducing inference times.
To summarize, the main contributions of our work are:
\begin{itemize}
    \setlength\itemsep{0em}
    \item a novel 3D semantic scene completion pipeline using an occupancy grid, %
    \item a lightweight architecture with mix of 2D/3D convolutions leading significantly less parameters,
    \item a modular multiscale pipeline which allows coarser inference at very high speed,
    \item state of the art performance on SemanticKITTI~\cite{Behley2019SemanticKITTIAD} and better performance on completion.
\end{itemize}

\section{Related Works}
\label{sec:related}

To process 3D data such as point-clouds, some use bird-eye-view \cite{chen2017multi} or 2D spherical \cite{Milioto2019RangeNetF} projection. Still, the common strategy relies either on point \cite{Qi2017PointNetDH} or voxel \cite{Riegler2017OctNetFusionLD} networks. 
The inherent limitation of voxel representation is the staggering memory requirement due to empty voxels, which led to optimized structures \cite{Riegler2017OctNetFusionLD} or use of sparse convolutions \cite{Graham20183DSS} to prevent dilation of the data manifold. \\
When it comes from real sensing, 3D data is inherently sparse (e.g. LiDAR scan, stereo, etc.) and its densification was initially framed as a \textit{completion or reconstruction} task. Recent works though, also assign semantic labels to their output subsequently referring to this as \textit{semantic completion}.

\paragraph{3D completion \& reconstruction.}
First works approximate missing data as a set of primitive local surfaces either from the data structure \cite{Thrun2005ShapeFS, Satkin20133DNNVI, Dai2017ShapeCU,Thrun2005ShapeFS} or using truncated Signed Distance Functions (TSDFs) \cite{Curless1996AVM, Newcombe2011KinectFusionRD, Roldo20193DSR}, while others use continuous energy minimization \cite{Kazhdan2006PoissonSR}.

More recently, learning methods boosted the completion of occluded and unseen regions \cite{Firman2016StructuredPO, Geiger2015Joint3O, Zimmermann2017LearningFA}. In \cite{Kim20133DSU}, voxel labels are predicted using a Conditional Random Field (CRF), while \cite{Dai2017ShapeCU} uses 3D-Encoder-Predictor to correlate observed scene with \textit{a priori} known 3D shapes. 
A few works also benefit from Signed Distance Functions representation as it provides richer gradient flow \cite{Park2019DeepSDFLC,Riegler2017OctNetFusionLD,Dai2019SGNNSG}. 
For memory reason, \cite{Dai2019SGNNSG} uses TSDFs input with sparse encoder and partially dense decoder to propagate features in unknown regions. 
While this effectively reduces memory, because TSDFs are denser than occupancy grids, we argue it would require a bigger and denser decoder, thus annihilating the benefit of any sparse encoding.
Other end-to-end completion networks were also proposed in \cite{Dai2017ScanNetR3, Chang2017Matterport3DLF}.
Despite discretization, we preferred a voxel-based implementation due to size limitations of point-based networks, even if there are promising object completion results~\cite{Yuan2018PCNPC}.

\paragraph{3D Semantic Scene Completion.}

SSCNet \cite{Song2017SemanticSC} was the first work to combine semantic segmentation and scene completion end-to-end with 3D CNNs. Further works also use additional RGB data by projecting or fusing semantic features from an image network \cite{Garbade2019TwoS3, Liu2018SeeAT, Li2019RGBDBD, Liu20203DGR}. An alternative to 3D data only is to encode LiDAR scans as spherical projection \cite{Milioto2019RangeNetF}, which enriches neighboring information \cite{Behley2019SemanticKITTIAD, Guo2018ViewVolumeNF}. While this boosts performance, it also increases the network complexity and subsequently the inference time. 
Generative Adversarial Networks (GANs) have also been proposed to enforce realistic outputs \cite{Wang2018AdversarialSS, Chen20193DSS} but are harder to train.
To lower memory consumption with the preferred voxelized representations, Spatial Group Convolutions (SGC) \cite{Zhang2018EfficientSS} divide input into groups for efficient processing at the cost of small performance drops.\\

Different from the literature, we rely solely on 3D voxelized occupancy data avoiding any preprocessing (as for TSDFs, SDFs, etc.), and propose a lightweight architecture with additional multiscale capability.

\begin{figure*}
    \centering
    \includegraphics[width=1.0\linewidth]{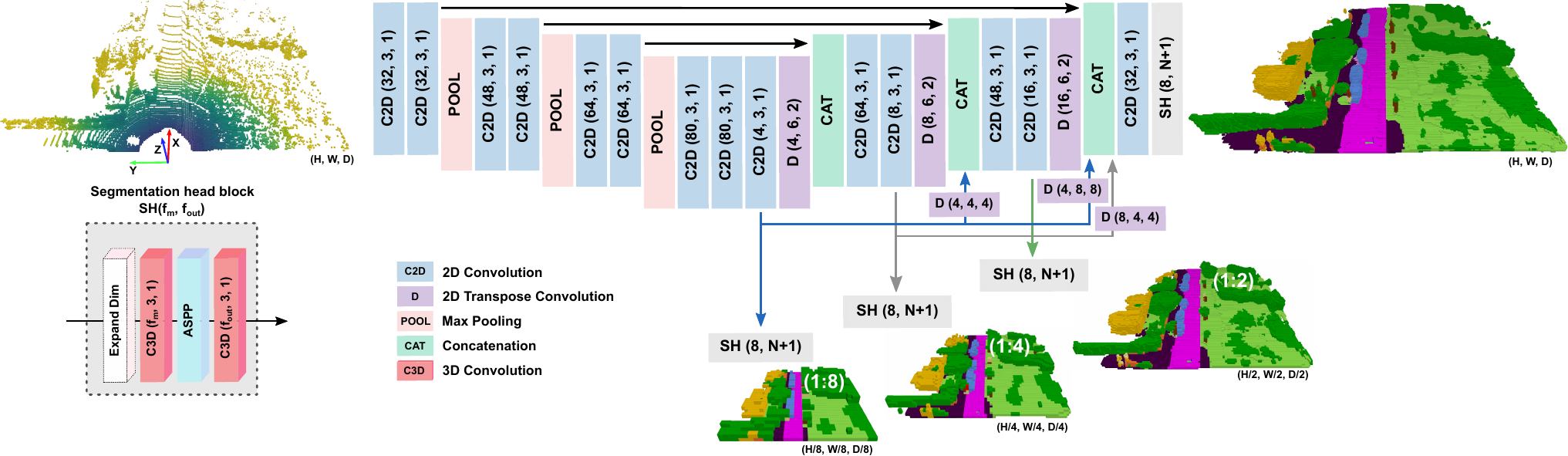}
    \caption{\textbf{LMSCNet: Lightweight Multiscale Semantic Completion Network.} Our pipeline uses a UNet architecture with 2D backbone convolutions (in blue) and 3D segmentation heads (in gray) to perform 3D semantic segmentation and completion at different scales, while preserving low complexity. Convolution parameters are shown as: (number of filters, kernel size and stride). Notice that we intentionally lower the 2D features dimension and use Atrous 3D convolutions (ASPP blocks from~\cite{Liu2018SeeAT}) to preserve low inference complexity.}
    \label{fig:architecture}
\end{figure*}

\section{Method}
\label{sec:method}

We tackle the problem of dense 3D semantic completion where the task is to assign a semantic label to each individual voxel. 
Given a sparse 3D voxel grid, the goal is to predict the 3D semantic scene representation, where each voxel is being assigned a semantic label $C = [c_{0}, c_{1}, \ldots, c_{N}]$, where $N$ is the number of semantic classes and $c_{0}$ stands for free voxels. 

Our architecture, coined LMSCNet and shown in Fig.~\ref{fig:architecture}, uses a lightweight UNet style architecture to predict 3D semantic completion at multiple scales, allowing fast coarse inference, beneficial for mobile robotics applications. 
Instead of greedy 3D convolutions, we mostly employ 2D convolutions along the height axis; similar to a bird-eye view.
In the following we detail our custom lightweight 2D/3D architecture (Sec.~\ref{sec:method-arch}), the multiscale reconstruction (Sec.~\ref{sec:method-mscale}), and the overall training pipeline (Sec.~\ref{sec:method-trainpipeline}).

\subsection{Lightweight multiscale 2D/3D architecture} \label{sec:method-arch}

To infer a dense output from the sparse input voxel grid, we use a standard encoder-decoder UNet architecture with 4 levels, thus learning features at decreasing resolutions. At each level, a series of convolution operations is applied followed by a pooling; downscaling the resolution size by 2. 
The reduction of spatial dimensions in UNets is beneficial for semantic tasks as it subsequently increases the kernels field-of-view at no cost. 
Note that dilated convolutions (a.k.a `atrous') with increasing dilation rates cannot be used in the encoder due to the sparse input nature. Though dense convolutions in the encoder imply a dilation of the input manifold~\cite{Graham20183DSS}, we argue this is beneficial for 3D semantic completion, given the sparse$\mapsto$dense nature of the task.

\paragraph{2D backbone.} To preserve a lightweight architecture, we use 2D convolutions along the X,Y dimensions, thus turning the height dimension (Z) into a feature dimension. Notice that we directly process 3D data in contrast to other 2D/3D works that rely on 2.5D data (e.g. depth \cite{Guo2018ViewVolumeNF, Liu2018SeeAT}, bird-eye view~\cite{chen2017multi}). While using 2D convolutions implies loosing 3D spatial connexity, it also enables significantly lighter operations. To further reduce the memory requirements, we keep a minimum number of features in each convolution layer.
Along with the standard skip connections, we also enhance information flow in the decoder by concatenating the output of each level to all lower levels. Technically, we upsample coarse feature maps learning ad-hoc deconvolution before concatenation to lower levels, which is shown with purple deconv blocks along blue and gray arrows in Fig.~\ref{fig:architecture}.
Intuitively, this enables our network to use high level features from coarser resolutions, and thus enhancing the spatial contextual information. 

\paragraph{3D segmentation head.}
Different from other works handling point cloud as bird-eye-view, the task of 3D semantic completion actually requires to retrieve the 3rd dimension ``lost'' with 2D convolutions. In other words, while 2D CNNs output 3D features maps, our decoder must output 4D tensor; the last dimension being the semantic class-wise probability distribution.\\
To address this, we introduce 3D segmentation heads depicted as gray blocks in Fig.~\ref{fig:architecture}. The heads use a series of dense and dilated convolutions. The latter, in the form of Atrous Spatial Pyramid Pooling (aka ASPP~\cite{Chen2018DeepLabSI, Liu2018SeeAT}), is beneficial to fuse information from different receptive fields thanks to the convolutions with increasing dilation rates (here [1, 2 and 3]). Note that dilated convolutions, though light and powerful, are not appropriate for sparse inputs and, as such, cannot be used in the encoder.
In our segmentation head, the benefit of preceding ASPP with dense 3D convolutions is dual: a) to further densify the feature maps, b) to ward off features from the segmentation heads and the backbone features. This last property is required to enable multiscale capacity, which we now describe.

\paragraph{Multiscale completion.} \label{sec:method-mscale}

\begin{figure}
    \centering
    \includegraphics[width=1\linewidth]{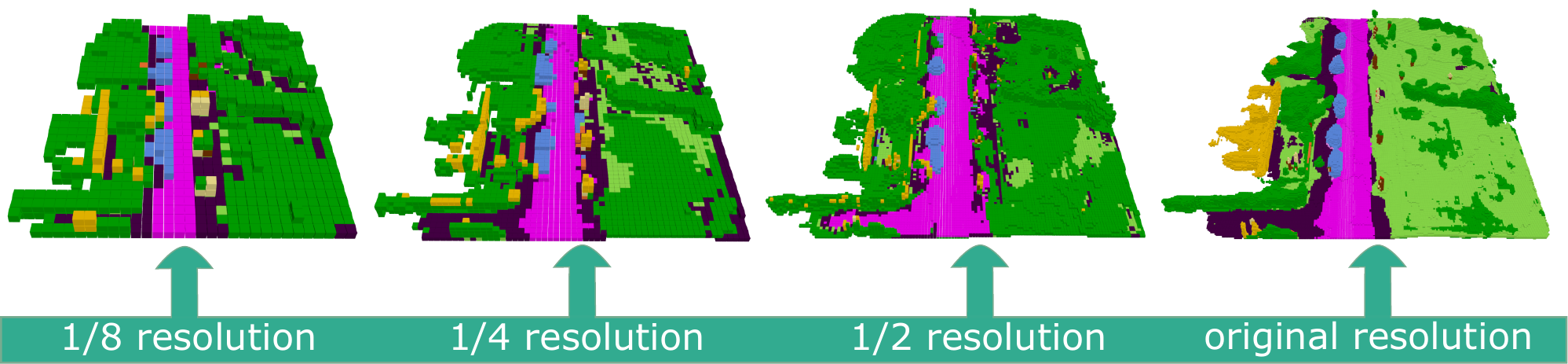}
    \caption{Our pipeline enables multiscale reconstruction. To supervise coarser representation, we use majority vote pooling from the original resolution ground truth.}
    \label{fig:mscale_pipeline}
\end{figure}

In the same vein as~\cite{Zhang2018EfficientSS, Dai2019SGNNSG}, we aim to output multiscale completion to enable both coarse scene representation and faster scene completion at lower resolution -- beneficial for mobile robotics applications.
We subsequently attach a 3D segmentation head after each level of the 2D UNet architecture, thus providing outputs at input relative scale of $\frac{1}{2^l}\ \forall\ l \in \{0,1,2,3\}$. A sample output is shown in Fig.~\ref{fig:mscale_pipeline}. 
As already mentioned, we noticed experimentally the importance of separating the segmentation features from the main features of the 2D backbone,  which again justifies the additional 3D convolutions in the segmentation head. The main interests of our multiscale architecture is that it infers semantic scene completion at a desired scale as needed, reducing the computation and memory requirements. This is further analyzed in Sec.~\ref{sec:exp-perf-ssc}.

\subsection{Training strategy} \label{sec:method-trainpipeline}

We train our LMSCNet from scratch in a standard end-to-end fashion from pairs of sparse input voxel ($x$) and semi-dense semantically labeled voxel grid ($\hat{y}$).
It is important to note that in a real setup, a dense ground truth is impractical for scene completion, due to occlusions and sensor field-of-view limitations. 
As such, the ground truth $\hat{y}$ is \textit{also} sparse and encoded with N+2 classes ($N$ semantic classes, 1 free class, 1 unknown). Similar to others \cite{Song2017SemanticSC, Liu2018SeeAT, Garbade2019TwoS3} we use a sparse loss strategy, back propagating the gradient only where ground truth is known.

\noindent{}For each scale $l$, we train with a cross-entropy loss defined as
\begin{equation}\label{eq:level_loss}
\mathcal{L}_{l} = - \sum_{c=0}^{N} w_{c}\hat{y}_{i, c}\log \left( \frac{e^{y_{i,c}}}{\textstyle \sum_{c^{\prime}}^{N} e^{y_{i,c^{\prime}}}} \right) \,,
\end{equation}
where $y$ is the network output, $i$ a voxel index, and $\hat{y}_{i, c}$ a one-hot vector (i.e. $\hat{y}_{i, c}=1$ if voxel $i$ is labeled class $c$, otherwise $\hat{y}_{i, c}=0$). 
Note that semantic tasks are by nature highly class-imbalanced problems. This is especially true in outdoor settings, which causes the prevalence of classes like road or vegetation. We account for the class-imbalance nature in Eq.~\ref{eq:level_loss} by weighting each class loss according to the inverse of the class-frequency $f_{c}$ as in \cite{Milioto2019RangeNetF}, thus using $w_{c} = \frac{1}{\log{\left( f_{c} + \epsilon \right)}}$ (with $\epsilon \ll 1$).
Finally, the complete network loss is a weighted sum of all level losses\footnote{In Eq.~\ref{eq:final_loss}, losses from heterogeneous resolutions can be summed due to the ad-hoc normalization in Eq.~\ref{eq:level_loss}} and writes:
\begin{equation}\label{eq:final_loss} 
\mathcal{L} = \sum^{3}_{l=0}\alpha_{l} \mathcal{L}_{l} \,,
\end{equation}

\noindent where $\alpha_l$ is the per-level loss weight, written for generality, though we use \mbox{$\alpha_l = 1, \forall l$} which works well and preserves multiscale capacity. Note that some of our choices were guided by faster training or inference speed. For example, unlike \cite{Zhang2018EfficientSS, Zhang2019CascadedCP, Dourado2019EdgeNetSS, Dai2019SGNNSG, Song2017SemanticSC} we avoid using Truncated Signed Distance Function variants (TSDF) that require a greedy computation time and was found to be of little benefit \cite{Garbade2019TwoS3, Behley2019SemanticKITTIAD}. We also tried to encode input as N+2 classes, that is with \textit{unknown} class, but we noticed little improvement -- if any -- at the cost of a large pre-processing time for ray casting.

\begin{table*}
	\scriptsize
	\setlength{\tabcolsep}{0.0035\linewidth}
	\newcommand{\classfreq}[1]{{~\tiny(\semkitfreq{#1}\%)}}  %
	\centering
	\begin{tabular}{l|c c c|c c c c c c c c c c c c c c c c c c c|c}
		\toprule
		& \multicolumn{3}{c|}{scene completion} & \multicolumn{20}{c}{semantic scene completion} \\
		Approach 
		& \rotatebox{90}{precision}
		& \rotatebox{90}{recall}
		& \rotatebox{90}{IoU}
		& \rotatebox{90}{\textcolor{road}{$\blacksquare$} road\classfreq{road}} 
		& \rotatebox{90}{\textcolor{sidewalk}{$\blacksquare$} sidewalk\classfreq{sidewalk}}
		& \rotatebox{90}{\textcolor{parking}{$\blacksquare$} parking\classfreq{parking}} 
		& \rotatebox{90}{\textcolor{other-ground}{$\blacksquare$} other-ground\classfreq{otherground}} 
		& \rotatebox{90}{\textcolor{building}{$\blacksquare$} building\classfreq{building}} 
		& \rotatebox{90}{\textcolor{car}{$\blacksquare$} car\classfreq{car}} 
		& \rotatebox{90}{\textcolor{truck}{$\blacksquare$} truck\classfreq{truck}} 
		& \rotatebox{90}{\textcolor{bicycle}{$\blacksquare$} bicycle\classfreq{bicycle}} 
		& \rotatebox{90}{\textcolor{motorcycle}{$\blacksquare$} motorcycle\classfreq{motorcycle}} 
		& \rotatebox{90}{\textcolor{other-vehicle}{$\blacksquare$} other-vehicle\classfreq{othervehicle}} 
		& \rotatebox{90}{\textcolor{vegetation}{$\blacksquare$} vegetation\classfreq{vegetation}} 
		& \rotatebox{90}{\textcolor{trunk}{$\blacksquare$} trunk\classfreq{trunk}} 
		& \rotatebox{90}{\textcolor{terrain}{$\blacksquare$} terrain\classfreq{terrain}} 
		& \rotatebox{90}{\textcolor{person}{$\blacksquare$} person\classfreq{person}} 
		& \rotatebox{90}{\textcolor{bicyclist}{$\blacksquare$} bicyclist\classfreq{bicyclist}} 
		& \rotatebox{90}{\textcolor{motorcyclist}{$\blacksquare$} motorcyclist\classfreq{motorcyclist}} 
		& \rotatebox{90}{\textcolor{fence}{$\blacksquare$} fence\classfreq{fence}} 
		& \rotatebox{90}{\textcolor{pole}{$\blacksquare$} pole\classfreq{pole}} 
		& \rotatebox{90}{\textcolor{traffic-sign}{$\blacksquare$} traffic-sign\classfreq{trafficsign}} 
		& \rotatebox{90}{mIoU}  \\
		\midrule
		SSCNet~\cite{Song2017SemanticSC} & 31.71 & 83.40 & 29.83 & 27.55 & 16.99 & 15.60 & 6.04 & 20.88 & 10.35 & 1.79 & 0 & 0 & 0.11 & 25.77 & 11.88 & 18.16 & 0 & 0 & 0 & 14.40 & 7.90 & 3.67 & 9.53  \\ %
		*SSCNet-full~\cite{Song2017SemanticSC} & 59.64 & 75.52 & 49.98 & 51.15 & 30.76 & 27.12 & 6.44 & 34.53 & 24.26 & 1.18 & \textbf{0.54} & \textbf{0.78} & \textbf{4.34} & 35.25 & 18.17 & 29.01 & 0.25 & 0.25 & \textbf{0.03} & 19.87 & 13.10 & 6.73 & 16.14  \\ %
		TS3D~\cite{Garbade2019TwoS3} & 31.58 & 84.18 & 29.81 & 28.00 & 16.98 & 15.65 & 4.86 & 23.19 & 10.72 & 2.39 & 0 & 0 & 0.19 & 24.73 & 12.46 & 18.32 & 0.03 & 0.05 & 0 & 13.23 & 6.98 & 3.52 & 9.54 \\ %
		TS3D+DNet~\cite{Behley2019SemanticKITTIAD} & 25.85 & \textbf{88.25} & 24.99 & 27.53 & 18.51 & 18.89 & \textbf{6.58} & 22.05 & 8.04 & 2.19 & 0.08 & 0.02 & 3.96 & 19.48 & 12.85 & 20.22 & \textbf{2.33} & \textbf{0.61} & 0.01 & 15.79 & 7.57 & \textbf{6.99} & 10.19 \\ %
		TS3D+DNet+SATNet~\cite{Behley2019SemanticKITTIAD} & 80.52 & 57.65 & 50.60 & 62.20 & 31.57 & 23.29 & 6.46 & 34.12 & 30.70 & \textbf{4.85} & 0 & 0 & 0.07 & 40.12 & \textbf{21.88} & \textbf{33.09} & 0 & 0 & 0 & \textbf{24.05} & \textbf{16.89} & 6.94 & \textbf{17.70} \\ %
		\midrule
		LMSCNet (ours) & 77.11 & 66.19 & 55.32 & 64.04& 33.12 & 24.91& 3.22 & \textbf{38.67} & 29.48 & 2.53 & 0 & 0 & 0.11 & 40.53 & 18.97 & 30.77 & 0 & 0 & 0 & 20.52 & 15.72 & 0.54 & 17.01 \\
		LMSCNet-singlescale (ours) &  \textbf{81.55} & 65.07 & \textbf{56.72} & \textbf{64.80} & \textbf{34.68} & \textbf{29.02} & 4.62 & 38.08 & \textbf{30.89} & 1.47 & 0 & 0 & 0.81 & \textbf{41.31} & 19.89 & 32.05 & 0 & 0 & 0 & 21.32 & 15.01 & 0.84 & 17.62 \\
		\bottomrule
	\end{tabular}\\
	{\scriptsize * Own implementation.}
	\caption{Comparison of published methods on the official SemanticKITTI~\cite{Behley2019SemanticKITTIAD} benchmark. Despite light mixed 2D/3D reasoning, our network performs 2nd on the semantic metrics (mIoU), outdistanced by the more complex TS3D+DNet+SATNet also twice slower than us. On the completion metrics (IoU), we perform 1st with a comfortable margin. The last two rows show that LMSCNet is better in its single scale version (\textit{LMSCNet-singlescale}), though this comes at the cost of loosing multiscale capacity. Except for SSCNet-full, all results originate from~\cite{Behley2019SemanticKITTIAD}.} 
	\label{table:net_reults}
\end{table*}

\begin{figure*}
	\centering
	\scriptsize
	\setlength{\tabcolsep}{0.017\linewidth}
	\renewcommand{\arraystretch}{0.8}
	\begin{tabular}{cccc}
		Input & SSCNet-full~\cite{Song2017SemanticSC} & LMSCNet (ours) & Ground Truth\\
		
		\includegraphics[width=0.4\columnwidth]{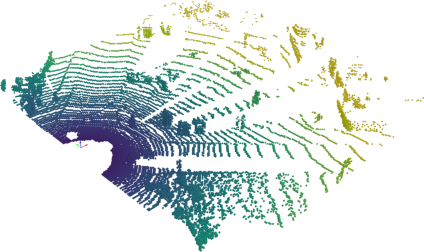} & 
		\includegraphics[width=0.4\columnwidth]{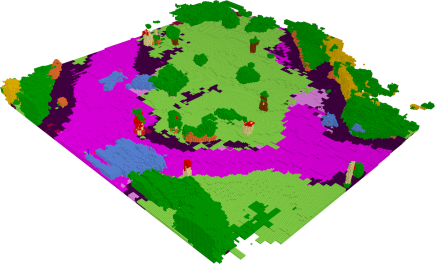} &
		\includegraphics[width=0.4\columnwidth]{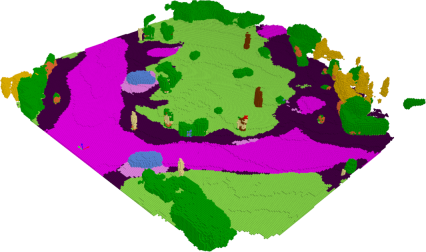} &
		\includegraphics[width=0.4\columnwidth]{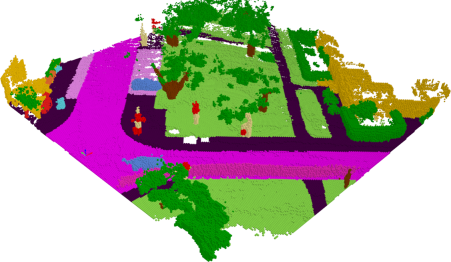} \\
		
		\includegraphics[width=0.4\columnwidth]{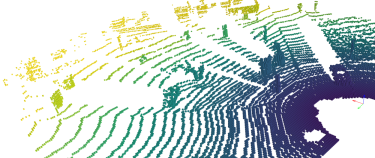} & 
		\includegraphics[width=0.4\columnwidth]{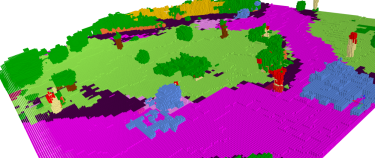} &
		\includegraphics[width=0.4\columnwidth]{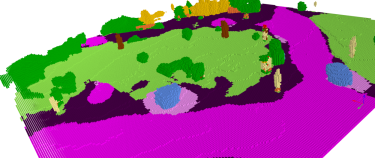} &
		{\includegraphics[width=0.4\columnwidth]{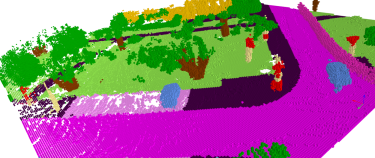}} \vspace{0.5cm}\\

		\includegraphics[width=0.4\columnwidth]{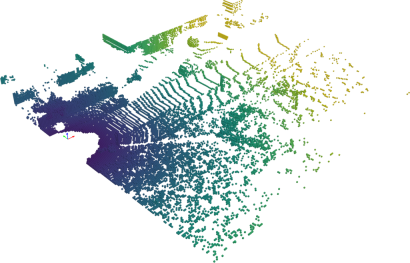} & 
		\includegraphics[width=0.4\columnwidth]{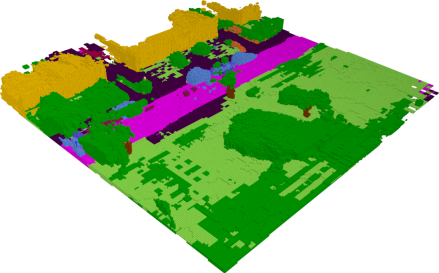} &
		\includegraphics[width=0.4\columnwidth]{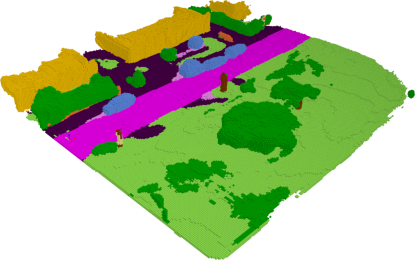} &
		\includegraphics[width=0.4\columnwidth]{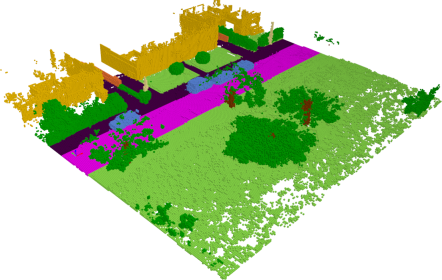} \\
		
		\includegraphics[width=0.4\columnwidth]{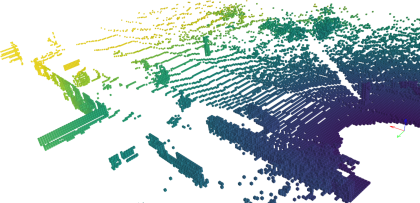} & 
		\includegraphics[width=0.4\columnwidth]{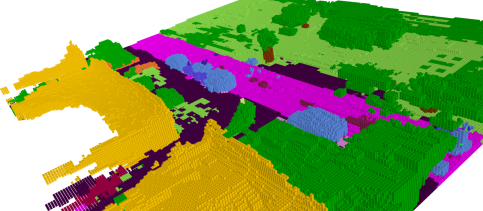} &
		\includegraphics[width=0.4\columnwidth]{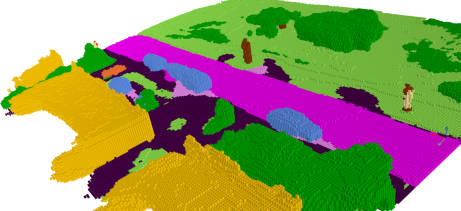} &
		{\includegraphics[width=0.4\columnwidth]{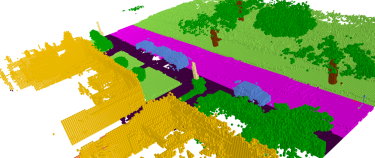}} \vspace{0.5cm}\\

		\includegraphics[width=0.4\columnwidth]{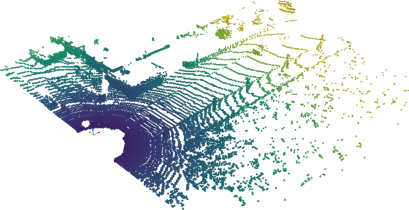} & 
		\includegraphics[width=0.4\columnwidth]{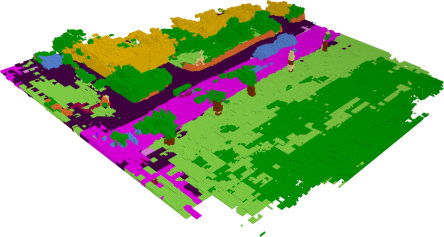} &
		\includegraphics[width=0.4\columnwidth]{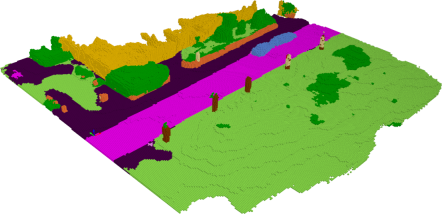} &
		\includegraphics[width=0.4\columnwidth]{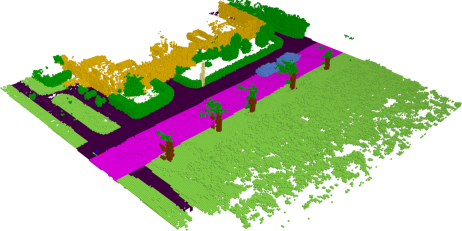} \\
		
		\includegraphics[width=0.4\columnwidth]{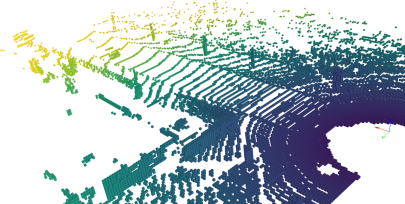} & 
		\includegraphics[width=0.4\columnwidth]{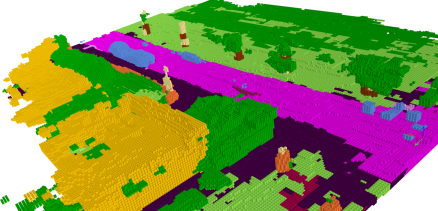} &
		\includegraphics[width=0.4\columnwidth]{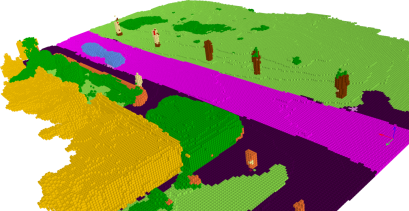} &
		{\includegraphics[width=0.4\columnwidth]{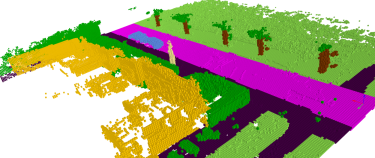}} \vspace{0.5cm}\\

		\includegraphics[width=0.4\columnwidth]{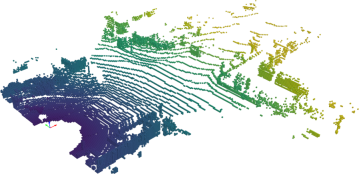} & 
		\includegraphics[width=0.4\columnwidth]{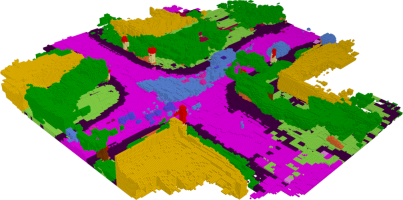} &
		\includegraphics[width=0.4\columnwidth]{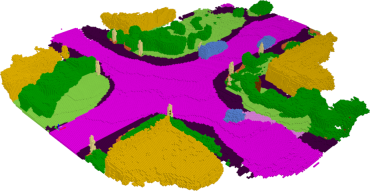} &
		\includegraphics[width=0.4\columnwidth]{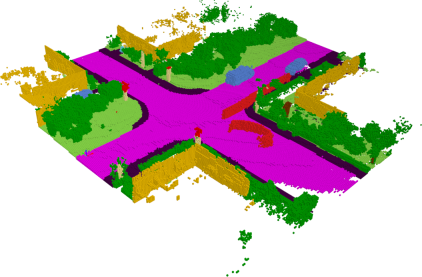} \\
		
		\includegraphics[width=0.4\columnwidth]{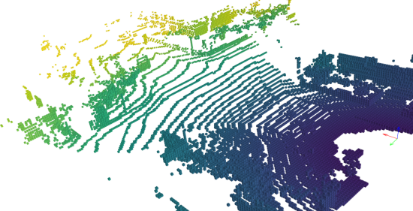} & 
		\includegraphics[width=0.4\columnwidth]{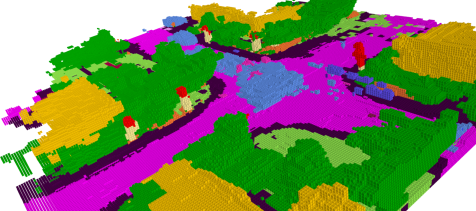} &
		\includegraphics[width=0.4\columnwidth]{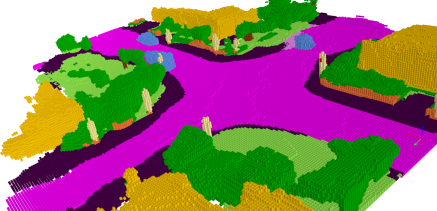} &
		{\includegraphics[width=0.4\columnwidth]{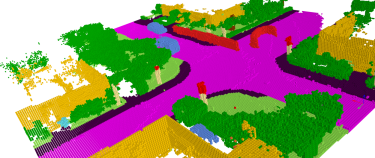}} \\
		
		\multicolumn{4}{c}{\includegraphics[width=1.7\columnwidth]{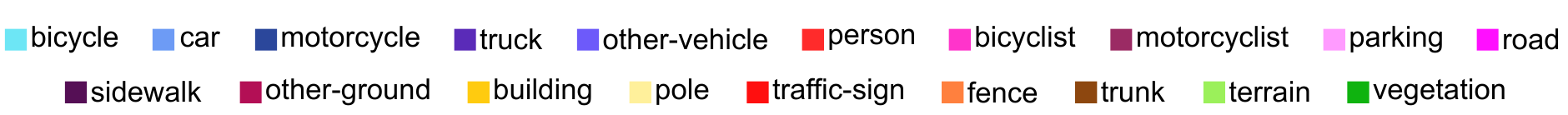}}
		
	\end{tabular}
	\caption{Qualitative 3D semantic completion at full size on the SemanticKITTI~\cite{Behley2019SemanticKITTIAD} validation set. Each pairs of rows show a single scene with different viewpoints. Compared to SSCNet-full~\cite{Song2017SemanticSC}, our LMSCNet provides smoother semantics labels and is capable of retrieving finer details. This is evident when looking at the cars (rows 7-8) or the trees (rows 5-6).}
	\label{fig:qualitative_comparison}
\end{figure*} 
\begin{figure}
	\centering
	\scriptsize
	\setlength{\tabcolsep}{0.005\linewidth}
	\renewcommand{\arraystretch}{0.7}
	\begin{tabular}{ccc}
	    Input & SSCNet-full~\cite{Song2017SemanticSC} & LMSCNet (ours) \\ 
		\includegraphics[width=0.32\linewidth]{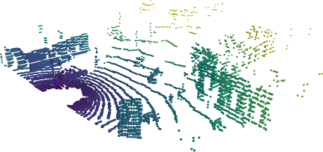}&\includegraphics[width=0.32\columnwidth]{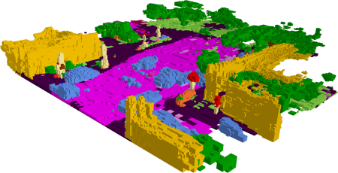}&\includegraphics[width=0.32\columnwidth]{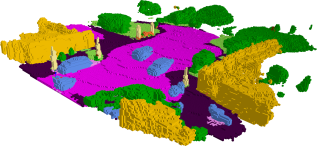} \\
	\end{tabular}
	\caption{Inference results on nuScenes~\cite{Caesar2019nuScenesAM} with 32-layers LiDAR, while being trained on 64-layers SemanticKITTI. Our method performs well with sharp scene labeling, despite the change of input density.}
	\label{fig:nuScenes_qualitative}
\end{figure}

\section{Experiments} \label{sec:experiments}
\label{sec:exp_metrics}
We evaluate our LMSCNet method by training on the recent semantic scene completion benchmark SemanticKITTI~\cite{Behley2019SemanticKITTIAD} providing 3D voxel grids from semantically labeled scans of HDL-64E rotating LiDAR in outdoor urban scenes~\cite{geiger2012we}. 
In~\cite{Behley2019SemanticKITTIAD}, inputs are voxelized single scans, while the ground truth was obtained from the voxelized aggregation of successive registered scans. Grids are 256x256x32 with 0.2m voxel size, and it is important to note that input \textit{and} ground truth are sparse, with average density of 6.7\% and 65.8\%, respectively. 
We use standard mIoU as a semantic completion metric, measuring the intersection over union averaged over all classes (20 semantic classes + \textit{free}). Additionally, we consider completion metrics IoU, Precision, and Recall to provide a sense of the scene completion quality, regardless of the assigned semantic labels (i.e. considering the binary \textit{free}~/~\textit{occupied} setting). Note that completion is crucial for obstacle avoidance in mobile robotics.

\paragraph{Implementation details.} 
We train using the original train/val splits with 3834/815 grids~\cite{Behley2019SemanticKITTIAD}, adding x-y flipping augmentation for generalization.
Adam optimizer is used ($\beta_{1}=0.9$, $\beta_{2}=0.999$) with learning rate of 0.001 scaled by $0.98^{\text{epoch}}$. Training fits in a single 11GB GPU with batch size 4, taking around 48 hours to converge (~80 epochs).

\subsection{Performance}\label{sec:exp-perf}
In the following we report performance against four state-of-the-art methods: SSCNet~\cite{Song2017SemanticSC}, TS3D~\cite{Garbade2019TwoS3}, TS3D+DNet~\cite{Behley2019SemanticKITTIAD}, TS3D+DNet+SATNet~\cite{Behley2019SemanticKITTIAD}. Because SSCNet output is 4x downsampled, we also report performance using deconvolution to reach full input resolution, hereafter denoted SSCNet-full. We refer to the supplementary for details on the required architectures adjustments.\\
\noindent{}Hereafter, we denote our multiscale architecture as \mbox{LMSCNet}. We detail semantic completion performance and then demonstrate the speed and lightness of our architecture. %

\subsubsection{Semantic Scene Completion}\label{sec:exp-perf-ssc}

Performance on the SemanticKITTI benchmark~\cite{Behley2019SemanticKITTIAD} is reported in Tab.~\ref{table:net_reults} against all published methods and SSCNet-full. 
The evaluation was conducted on the official server (i.e. hidden test set) hence, with the full size ground truth.

Overall, we perform on par with the best methods, though 2nd on the semantic completion metric (mIoU). On the latter, TS3D+DNet+SATNet~\cite{Behley2019SemanticKITTIAD} is slightly better despite their significantly heavier and slower network. Note also that TS3D uses additional RGB input, and all TS3D+DNet use also LiDAR refraction intensity. Conversely, LMSCNet is more versatile as it uses only occupancy grid input.
Notice that the highly imbalanced class frequencies (shown in parenthesis in Tab.~\ref{table:net_reults}) also illustrate the task complexity. Specifically, we outperform others on the largest four classes but performs on par or lower on the others, which advocates for some improvement in our balancing strategy. 
On the completion metrics (IoU) our method outperforms all others by a comfortable margin. Again, completion is of high importance for practical mobile robotics applications. 

In addition to the multiscale proposal (LMSCNet), we also report LMSCNet-singlescale -- a variation of LMSCNet where we train with $\mathcal{L} = \mathcal{L}_0$ --, which logically performs a little better at full size though at the cost of loosing crucial multiscale capacity.

\paragraph{Qualitative performance.}

We compare qualitatively full size outputs of our LMSCNet and SSCNet-full in Fig.~\ref{fig:qualitative_comparison}, with views pairs from 4 scenes of the SemanticKITTI validation set\footnote{Note that SemanticKITTI benchmark (i.e. test set) does not provide any visual results. Hence, we omit TS3D baselines due to retraining complexities and their use of additional modalities (RGB or LiDAR intensity).}. 
At the rightmost, ground truth visualization also illustrates the sparse supervision complexity since holes are still visible.
Our method produces visually smoother semantic labels, easily noticeable in rows 5-8, and is able to reconstruct thin structures, like trees or cars (rows 6 or 7). 
For comprehensive analysis, we further test the same model (trained on 64-layer LiDAR) on the popular nuScenes dataset~\cite{Caesar2019nuScenesAM}, which has been registered using a 32-layers lidar sensor. Fig.~\ref{fig:nuScenes_qualitative} shows that our network better adjusts to the change of density and maintains the smoothness in the reconstruction. \vspace{-0.3 cm}

\paragraph{Multiscale performance.}
\begin{table}
		\setlength{\tabcolsep}{0.0038\linewidth}
		\centering
		\footnotesize
		\begin{tabular}{lccc}
			\toprule
			LMSCNet scale & IoU & mIoU  \\
			\midrule
			1:1 (full size) & 54.22 & 16.78 \\
			1:2 & 56.27 & 16.78 \\
			1:4 & 59.36 & 17.19 \\
			1:8 & 65.45 & 17.37 \\
			\bottomrule
	\end{tabular}
	\caption{LMSCNet multiscale semantic completion performance on SemanticKITTI validation set. We reach good performance at all levels, even better at the coarsest levels .\vspace{-0.5 cm}}
	\label{table:multiscale}
\end{table}
Tab.~\ref{table:multiscale} shows multiscale performance of our method on the SemanticKITTI validation set, where the scale is relative to the full size resolution~(level~0). From Sec.~\ref{sec:method-mscale}, scale at level $l$ is $\frac{1}{2^l}$. Ground truths at lower resolution were obtained from majority vote pooling of the full size ground truth.
From the above table, our architecture maintains a good performance in all resolutions, with best performance logically reached at the lowest resolution (highest level). 
Qualitative multiscale completion is visible in Fig.~\ref{fig:mscale_pipeline}.
We argue that our architecture reaches multiscale capacity thanks to the disentanglement of segmentation features with our custom head. Additionally, at coarser resolution our network reaches very fast inference, which will be described in details in the following section.

\subsubsection{Architectures comparison.}\label{sec:exp-perf-arch}

\begin{table}[!h]
	\footnotesize
	\centering
	\setlength{\tabcolsep}{0.01\linewidth}
	\begin{tabular}{cccc}
		\toprule
		Method & Params (M) &  FLOPs (G) & FPS\\
		\midrule
		*SSCNet~\cite{Song2017SemanticSC}                    & 0.93 & 82.5 & 56.90 \\  
		*SSCNet-full~\cite{Song2017SemanticSC}             & 1.09 & 769.6 & 45.94 \\
		*TS3D~\cite{Garbade2019TwoS3}                        & 43.77 & 2016.7 & 9.79 \\  %
		*TS3D+DNet~\cite{Behley2019SemanticKITTIAD}          & 51.31 & 847.1 & 8.72\\ %
		*TS3D+DNet+SATNet~\cite{Behley2019SemanticKITTIAD}   & 50.57 & 905.2 & 1.27\\  %
		\midrule
		LMSCNet			                                 & 0.35 & 72.6 & 21.28\\
		LMSCNet (1:2)                                 & 0.32 & 13.7  & 126.38\\
		LMSCNet (1:4)                                 & 0.28 & 5.7  & 323.46\\
		LMSCNet (1:8)                                 & 0.24 & 4.4  & 372.24\\
		\bottomrule
	\end{tabular}\\
	{\scriptsize * Own implementation to compute network statistics.}
	\caption{Network statistics. Even at full resolution \mbox{LMSCNet} (ours) has significantly less parameters with lower FLOPs. On a speed basis, we are twice slower than SSCNet-full~\cite{Song2017SemanticSC} which performs worse than us (see Tab.~\ref{table:net_reults}). Still, our multiscale versions -- denoted \mbox{\textit{LMSCNet (1:x)}} -- enable very fast inference.}
	\label{table:net_stats}
\end{table}

\begin{figure}
	\centering
	\scriptsize
	\setlength{\tabcolsep}{0.005\linewidth}
	\renewcommand{\arraystretch}{0.7}
	\begin{tabular}{cc}
		\includegraphics[width=0.5\linewidth]{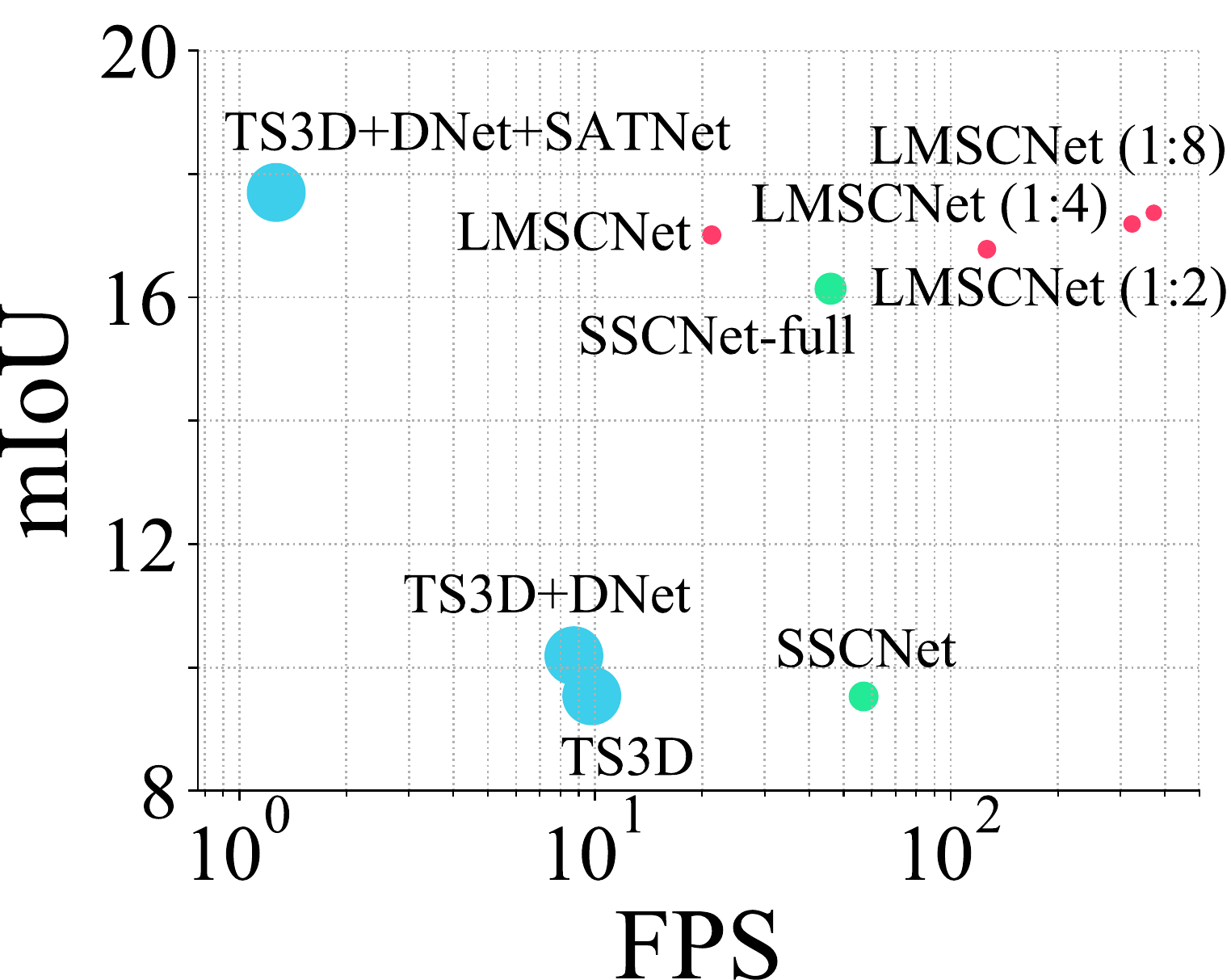}&\includegraphics[width=0.5\columnwidth]{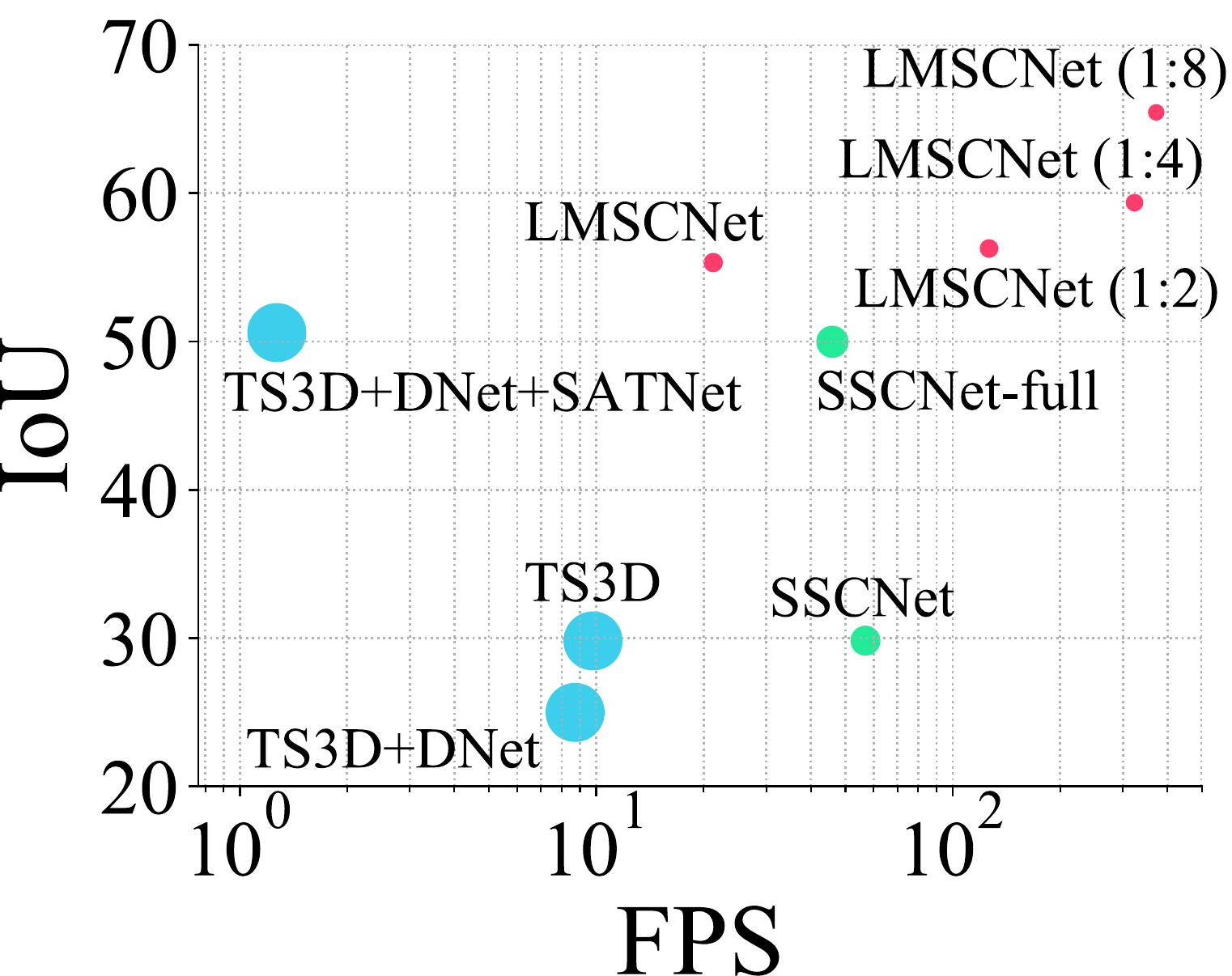}\\
	\end{tabular}\vspace{-1em}
	\caption{Architectures performance versus speed (markers are scaled with \# of parameters). Notice that \mbox{TS3D+DNet+SATNet} is the only better method on semantics (+0.69 mIoU) though less time performant (x17 slower) and worse on completion (-4.72 IoU).}
	\label{fig:scatter-perf-speed-params}
\end{figure} 

Tab.~\ref{table:net_stats} reports networks statistics for our architecture and all above mentioned baselines. From the latter, even at full size LMSCNet has significantly less parameters (0.35M) and lower computational cost for inference (72.6G FLOPs). Compared to any TS3D baselines it is at least an order of magnitude faster. However, SSCNet (original or full) is twice faster than LMSCNet, though with more parameters and worse performance (cf. Tab.~\ref{table:net_reults}). Since lighter models does not \textit{always} run faster due to the sequentiality of some operations on GPU, we conjecture the higher speed of SSCNet is caused by the lower number of convolutional operations compared to LMSCNet full scale (16 vs. 25).

In last rows of Tab.~\ref{table:net_stats}, we report statistics for coarser completion, removing \textit{unnecessary} parts of our network at inference. 
Lower resolution inference allows significant speedups in the processing, reaching 372 FPS at the highest scale -- 6x faster than SSCNet and ~300x faster than \mbox{TS3D+DNet+SATNet} --.
Fig. \ref{fig:scatter-perf-speed-params} illustrates the architectures performance versus speed. Notice that \textit{even at full scale} we provide a better speed-performance balance.
Because semantic completion is an application of high interest for mobile robotics, like autonomous driving, our lighter architecture is beneficial for embedded GPUs and enables coarse scene analysis at high speed.

\begin{figure}
	\centering
	\scriptsize
	\setlength{\tabcolsep}{0.01\linewidth}
	\renewcommand{\arraystretch}{0.7}
	\begin{tabular}{ccc}
		\includegraphics[height=1.75cm]{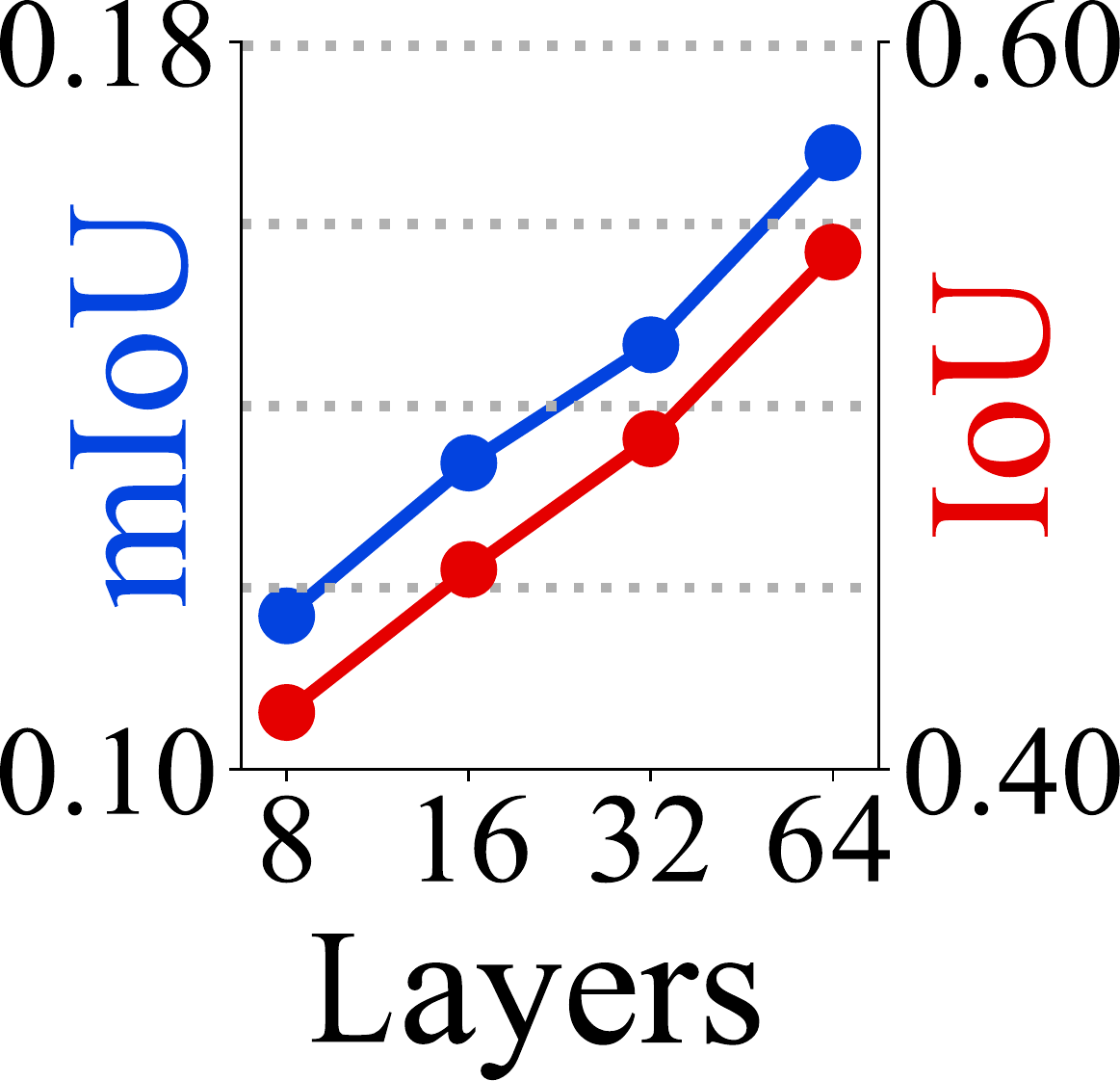}&\includegraphics[width=0.33\columnwidth]{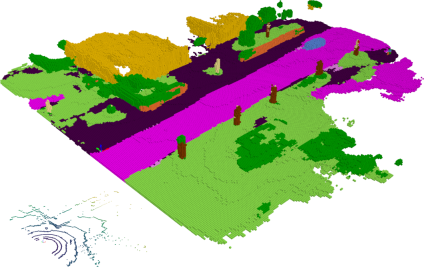}&\includegraphics[width=0.33\columnwidth]{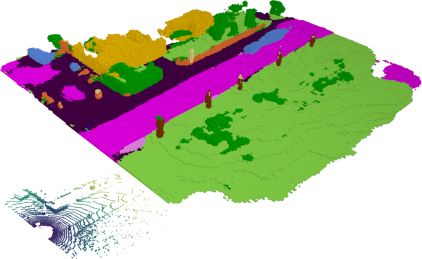}\\
		& 8 layers & 32 layers
	\end{tabular}
	\caption{Semantic scene completion results from simulated lower resolution LiDAR sensors (downsampled from 64 layers input). Even with only 8 layers input our LMSCNet correctly predicts the scene outline.}
	\label{fig:lidar_multires_qualitative}\label{fig:lidar_res_plot}
\end{figure}

\subsection{Ablation studies}
\label{sec:exp-ablation}

To study the benefit of our design choices, we conduct a series of ablation studies on SemanticKITTI validation set. This is done by modifying important blocks of our architecture and evaluating its performance. 

\paragraph{Influence of input resolution.}
We evaluate our robustness, by retrieving the original 64-layers KITTI scans used in SemanticKITTI and simulating 8/16/32 layers LiDARs with layers subsampling\footnote{Every 2nd, 4th and 8th layer are subsampled to simulate 32, 16 and 8 layer LiDARs, respectively. Unlike~\cite{Jaritz2018SparseAD}, note that data SemanticKITTI uses KITTI odometry set in which data is already untwisted.}, as in~\cite{Jaritz2018SparseAD}.

Fig.~\ref{fig:lidar_res_plot} shows quantitative and qualitative performance using simulated and original LiDAR. As expected, lower layers input deteriorate the performance, especially in areas far from the sensor location, but our network still performs reasonably well on semantics (mIoU) and completion (IoU). This is visible in the middle image, as 8 layers input (2.10\% density) is sufficient to retrieve the general outline of the scene.

\begin{table}
\centering
	\setlength{\tabcolsep}{0.01\linewidth}
		\centering
		\footnotesize
		\begin{tabular}{lccc}
			\toprule
			Method & IoU & mIoU  \\
			\midrule
			LMSCNet (ours) & 54.22 & 16.78 \\
			w/o Deconv & 52.79 &	15.64 \\
			w/o ASPP & 53.81 &	16.21 \\
			w/o Multiscale UNet & 53.54 &	16.22 \\
			\bottomrule
	\end{tabular}
\caption{Ablation study of our model design choices on the SemanticKITTI~\cite{Behley2019SemanticKITTIAD} validation set. } 
\label{table:ablation_table}
\end{table}
\paragraph{Deconv versus Upsampling.}

As we aimed to preserve a lightweight architecture, we tried to remove the parameters-greedy deconv layers from our network (cf. Sec.~\ref{fig:architecture}), replacing them with up-sampling layers. From Tab.~\ref{table:ablation_table}, performance \textit{without deconv} introduces a 1.43\% and 1.14\% performance drop for completion and semantic completion respectively, with only 3\% less parameters.

\paragraph{Dilated convolutions.}
We evaluate the benefit of dilated convolutions in the decoder by ablating ASPP blocks from the segmentation head (see Fig.~\ref{fig:architecture}). Tab.~\ref{table:ablation_table} indicates that mIoU drops by 0.41\% without ASPP. We conjecture that the boost of ASSP results come from the increasing receptive fields of the inner dilated convolutions, providing richer features.

\paragraph{Multiscale UNet decoder.}
\begin{figure}
\centering
\subfloat[Vanilla UNet decoder]{\includegraphics[width=0.455\columnwidth]{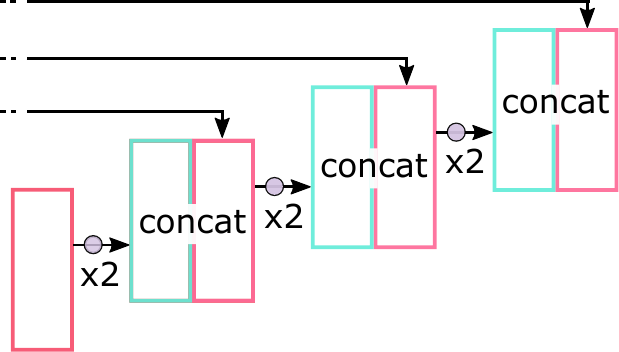}\label{fig:scheme_non_allf_concat}}
\hspace{0.02\columnwidth}
\subfloat[Multiscale UNet decoder]{\includegraphics[width=0.49\columnwidth]{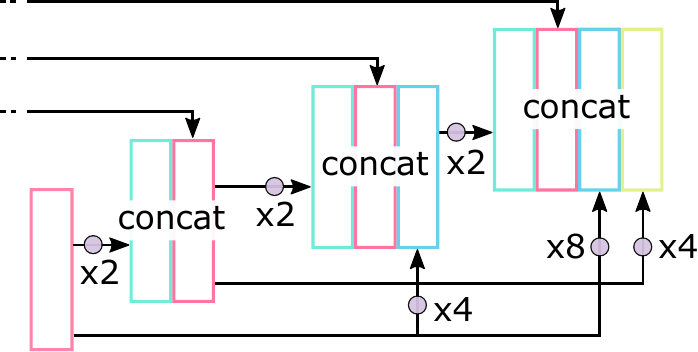}\label{fig:scheme_allf_concat}}
\caption{Decoders comparison. While Vanilla UNet decoder only considers features from the previous level \protect\subref{fig:scheme_non_allf_concat}, we instead use Multiscale UNet where all coarser levels enhance spatial contextual information \protect\subref{fig:scheme_allf_concat}. Circles show intermediary operations to reach required feature maps size.}
\label{fig:schemes_decoder_vanilla_vs_mscale}
\end{figure}

As illustrated in Fig.~\ref{fig:scheme_non_allf_concat}, unlike vanilla UNet decoder we concatenate the features at the end of each decoder level to all other levels. This is intended to aggregate multiscale features and should intuitively help considering coarser semantic features for fine resolutions.

We assess the benefit of our \textit{multiscale UNet} by evaluating \textit{Vanilla UNet} in the last row of Tab.~\ref{table:ablation_table}, which shows that our proposal boosts completion by 0.68\% and semantic completion by 0.56\%.

\section{Conclusion}
\label{sec:concl}

We proposed a novel method, coined LMSCNet, for 3D Semantic Scene Completion, which benefits from mixing 2D/3D convolutions to preserve lightweight architecture, while enabling inference at multiple scales. 
On the challenging SemanticKITTI benchmark, we perform on par with other methods on semantic completion with a much lighter architecture and at faster inference speed. For completion, we outperform the state of the art. Results show that loss of 3D spatial connexity caused by the 2D convolutions \textit{does not} impair performance. We attribute this to the uniform dimensional variance in used application (i.e. constant sensor viewpoint, urban outdoor scenes). We conjecture that data with higher variance in all directions would cause a higher impact. Also of interest for mobile robotics, our proposal is robust to much lower input density and our multiscale capacity enables scene completion for lower resolution at very high speed.

{\small
\bibliographystyle{ieee}
\bibliography{biblio}

\begin{thebibliography}{10}\itemsep=-1pt

\bibitem{Behley2019SemanticKITTIAD}
J.~Behley, M.~Garbade, A.~Milioto, J.~Quenzel, S.~Behnke, C.~Stachniss, and
  J.~Gall.
\newblock {SemanticKITTI}: A dataset for semantic scene understanding of
  {LiDAR} sequences.
\newblock {\em International Conference on Computer Vision (ICCV)}, pages
  9296--9306, 2019.

\bibitem{Berger2017ASO}
M.~Berger, A.~Tagliasacchi, L.~M. Seversky, P.~Alliez, G.~Guennebaud, J.~A.
  Levine, A.~Sharf, and C.~T. Silva.
\newblock A survey of surface reconstruction from point clouds.
\newblock {\em Comput. Graph. Forum}, 36:301--329, 2017.

\bibitem{Caesar2019nuScenesAM}
H.~Caesar, V.~Bankiti, A.~H. Lang, S.~Vora, V.~E. Liong, Q.~Xu, A.~Krishnan,
  Y.~Pan, G.~Baldan, and O.~Beijbom.
\newblock {nuScenes}: A multimodal dataset for autonomous driving.
\newblock {\em Conference on Computer Vision and Pattern Recognition (CVPR)},
  2020.

\bibitem{Chang2017Matterport3DLF}
A.~X. Chang, A.~Dai, T.~A. Funkhouser, M.~Halber, M.~Nie{\ss}ner, M.~Savva,
  S.~Song, A.~Zeng, and Y.~Zhang.
\newblock {Matterport3D}: Learning from {RGB-D} data in indoor environments.
\newblock {\em International Conference on 3D Vision (3DV)}, pages 667--676,
  2017.

\bibitem{Chen2018DeepLabSI}
L.-C. Chen, G.~Papandreou, I.~Kokkinos, K.~Murphy, and A.~L. Yuille.
\newblock {DeepLab}: Semantic image segmentation with {Deep Convolutional
  Nets}, {Atrous Convolution}, and {Fully Connected} {CRFs}.
\newblock {\em Transactions on Pattern Analysis and Machine Intelligence
  (PAMI)}, 40:834--848, 2018.

\bibitem{chen2017multi}
X.~Chen, H.~Ma, J.~Wan, B.~Li, and T.~Xia.
\newblock Multi-view {3D} object detection network for autonomous driving.
\newblock In {\em Conference on Computer Vision and Pattern Recognition
  (CVPR)}, pages 1907--1915, 2017.

\bibitem{Chen20193DSS}
Y.~Chen, M.~Garbade, and J.~Gall.
\newblock {3D} semantic scene completion from a single depth image using
  adversarial training.
\newblock In {\em International Conference on Image Processing (ICIP)}, pages
  1835--1839, 2019.

\bibitem{Curless1996AVM}
B.~Curless and M.~Levoy.
\newblock A volumetric method for building complex models from range images.
\newblock In {\em SIGGRAPH}, 1996.

\bibitem{Dai2017ScanNetR3}
A.~Dai, A.~X. Chang, M.~Savva, M.~Halber, T.~A. Funkhouser, and M.~Nie{\ss}ner.
\newblock {ScanNet}: Richly-annotated {3D} reconstructions of indoor scenes.
\newblock {\em Conference on Computer Vision and Pattern Recognition (CVPR)},
  pages 2432--2443, 2017.

\bibitem{Dai2019SGNNSG}
A.~Dai, C.~Diller, and M.~Nie{\ss}ner.
\newblock {SG-NN}: Sparse generative neural networks for self-supervised scene
  completion of {RGB-D} scans.
\newblock {\em Conference on Computer Vision and Pattern Recognition (CVPR)},
  2020.

\bibitem{Dai2017ShapeCU}
A.~Dai, C.~R. Qi, and M.~Nie{\ss}ner.
\newblock Shape completion using {3D}-encoder-predictor {CNNs} and shape
  synthesis.
\newblock {\em Conference on Computer Vision and Pattern Recognition (CVPR)},
  pages 6545--6554, 2017.

\bibitem{Dai2018ScanCompleteLS}
A.~Dai, D.~Ritchie, M.~Bokeloh, S.~Reed, J.~Sturm, and M.~Nie{\ss}ner.
\newblock {ScanComplete}: Large-scale scene completion and semantic
  segmentation for {3D} scans.
\newblock {\em Conference on Computer Vision and Pattern Recognition (CVPR)},
  pages 4578--4587, 2018.

\bibitem{Dourado2019EdgeNetSS}
A.~Dourado, T.~E. de~Campos, H.~S. Kim, and A.~Hilton.
\newblock {EdgeNet}: Semantic scene completion from {RGB-D} images.
\newblock {\em ArXiv}, abs/1908.02893, 2019.

\bibitem{Firman2016StructuredPO}
M.~Firman, O.~M. Aodha, S.~J. Julier, and G.~J. Brostow.
\newblock Structured prediction of unobserved voxels from a single depth image.
\newblock {\em Conference on Computer Vision and Pattern Recognition (CVPR)},
  pages 5431--5440, 2016.

\bibitem{Garbade2019TwoS3}
M.~Garbade, J.~Sawatzky, A.~Richard, and J.~Gall.
\newblock Two stream {3D} semantic scene completion.
\newblock {\em Conference on Computer Vision and Pattern Recognition Workshops
  (CVPRW)}, pages 416--425, 2019.

\bibitem{geiger2012we}
A.~Geiger, P.~Lenz, and R.~Urtasun.
\newblock Are we ready for autonomous driving? {The} {KIITI} vision benchmark
  suite.
\newblock In {\em Conference on Computer Vision and Pattern Recognition
  (CVPR)}, pages 3354--3361, 2012.

\bibitem{Geiger2015Joint3O}
A.~Geiger and C.~Wang.
\newblock Joint {3D} object and layout inference from a single {RGB-D} image.
\newblock In {\em German Conference on Pattern Recognition (GCPR)}, 2015.

\bibitem{Graham20183DSS}
B.~Graham, M.~Engelcke, and L.~van~der Maaten.
\newblock {3D} semantic segmentation with submanifold sparse convolutional
  networks.
\newblock {\em Conference on Computer Vision and Pattern Recognition (CVPR)},
  pages 9224--9232, 2018.

\bibitem{Guo2018ViewVolumeNF}
Y.-X. Guo and X.~Tong.
\newblock View-volume network for semantic scene completion from a single depth
  image.
\newblock In {\em International Joint Conference on Artificial Intelligence
  (IJCAI)}, 2018.

\bibitem{Jaritz2018SparseAD}
M.~Jaritz, R.~de~Charette, {\'E}.~Wirbel, X.~Perrotton, and F.~Nashashibi.
\newblock Sparse and dense data with {CNNs}: Depth completion and semantic
  segmentation.
\newblock In {\em International Conference on 3D Vision (3DV)}, pages 52--60,
  2018.

\bibitem{Kazhdan2006PoissonSR}
M.~M. Kazhdan, M.~Bolitho, and H.~Hoppe.
\newblock Poisson surface reconstruction.
\newblock In {\em Symposium on Geometry Processing (SGP)}, 2006.

\bibitem{Kim20133DSU}
B.-S. Kim, P.~Kohli, and S.~Savarese.
\newblock {3D} scene understanding by voxel{-CRF}.
\newblock {\em International Conference on Computer Vision (ICCV)}, pages
  1425--1432, 2013.

\bibitem{Kober2013ReinforcementLI}
J.~Kober, J.~A. Bagnell, and J.~Peters.
\newblock Reinforcement learning in robotics: A survey.
\newblock {\em International Journal of Robotics Research (IJRR)}, 32:1238 --
  1274, 2013.

\bibitem{Li2019RGBDBD}
J.~Li, Y.~Liu, D.~Gong, Q.~Shi, X.~Yuan, C.~Zhao, and I.~D. Reid.
\newblock {RGBD} based dimensional decomposition residual network for {3D}
  semantic scene completion.
\newblock {\em Conference on Computer Vision and Pattern Recognition (CVPR)},
  pages 7685--7694, 2019.

\bibitem{Li2009TowardsTS}
L.-J. Li, R.~Socher, and L.~Fei-Fei.
\newblock Towards total scene understanding: Classification, annotation and
  segmentation in an automatic framework.
\newblock {\em Conference on Computer Vision and Pattern Recognition (CVPR)},
  pages 2036--2043, 2009.

\bibitem{Liu2018SeeAT}
S.~Liu, Y.~Hu, Y.~Zeng, Q.~Tang, B.~Jin, Y.~Han, and X.~Li.
\newblock See and think: Disentangling semantic scene completion.
\newblock In {\em NeurIPS}, 2018.

\bibitem{Liu20203DGR}
Y.~W. Liu, J.~Li, Q.~Yan, X.~Yuan, C.-X. Zhao, I.~Reid, and C.~Cadena.
\newblock {3D} gated recurrent fusion for semantic scene completion.
\newblock {\em ArXiv}, abs/2002.07269, 2020.

\bibitem{Liu2019PointVoxelCF}
Z.~Liu, H.~Tang, Y.~Lin, and S.~Han.
\newblock Point-voxel {CNN} for efficient {3D} deep learning.
\newblock In {\em NeurIPS}, 2019.

\bibitem{Milioto2019RangeNetF}
A.~Milioto, I.~Vizzo, J.~Behley, and C.~Stachniss.
\newblock {RangeNet ++}: Fast and accurate {LiDAR} semantic segmentation.
\newblock {\em International Conference on Intelligent Robots and Systems
  (IROS)}, pages 4213--4220, 2019.

\bibitem{Newcombe2011KinectFusionRD}
R.~A. Newcombe, S.~Izadi, O.~Hilliges, D.~Molyneaux, D.~Kim, A.~J. Davison,
  P.~Kohli, J.~Shotton, S.~Hodges, and A.~W. Fitzgibbon.
\newblock {KinectFusion}: Real-time dense surface mapping and tracking.
\newblock {\em International Symposium on Mixed and Augmented Reality}, pages
  127--136, 2011.

\bibitem{Park2019DeepSDFLC}
J.~J. Park, P.~Florence, J.~Straub, R.~A. Newcombe, and S.~Lovegrove.
\newblock {DeepSDF}: Learning continuous signed distance functions for shape
  representation.
\newblock {\em Conference on Computer Vision and Pattern Recognition (CVPR)},
  pages 165--174, 2019.

\bibitem{Qi2017PointNetDH}
C.~R. Qi, L.~Yi, H.~Su, and L.~J. Guibas.
\newblock Pointnet++: Deep hierarchical feature learning on point sets in a
  metric space.
\newblock In {\em NeurIPS}, 2017.

\bibitem{Riegler2017OctNetFusionLD}
G.~Riegler, A.~O. Ulusoy, H.~Bischof, and A.~Geiger.
\newblock {OctNetFusion}: Learning depth fusion from data.
\newblock In {\em International Conference on 3D Vision (3DV)}, pages 57--66,
  2017.

\bibitem{Roldo20193DSR}
L.~Rold{\~a}o, R.~de~Charette, and A.~Verroust-Blondet.
\newblock {3D} surface reconstruction from voxel-based {LiDAR} data.
\newblock {\em Intelligent Transportation Systems Conference (ITSC)}, pages
  2681--2686, 2019.

\bibitem{Satkin20133DNNVI}
S.~Satkin and M.~Hebert.
\newblock {3DNN}: Viewpoint invariant {3D} geometry matching for scene
  understanding.
\newblock {\em International Conference on Computer Vision (ICCV)}, pages
  1873--1880, 2013.

\bibitem{Song2017SemanticSC}
S.~Song, F.~Yu, A.~Zeng, A.~X. Chang, M.~Savva, and T.~A. Funkhouser.
\newblock Semantic scene completion from a single depth image.
\newblock {\em Conference on Computer Vision and Pattern Recognition (CVPR)},
  pages 190--198, 2017.

\bibitem{Thrun2005ShapeFS}
S.~Thrun and B.~Wegbreit.
\newblock Shape from symmetry.
\newblock {\em International Conference on Computer Vision (ICCV)},
  2:1824--1831 Vol. 2, 2005.

\bibitem{Krevelen2010ASO}
D.~W.~F. van Krevelen and R.~Poelman.
\newblock A survey of augmented reality technologies, applications and
  limitations.
\newblock {\em International Journal of Virtual Reality (IJVR)}, 9:1--20, 2010.

\bibitem{Wang2018AdversarialSS}
Y.~Wang, D.~J. Tan, N.~Navab, and F.~Tombari.
\newblock Adversarial semantic scene completion from a single depth image.
\newblock In {\em International Conference on 3D Vision (3DV)}, pages 426--434,
  2018.

\bibitem{Yuan2018PCNPC}
W.~Yuan, T.~Khot, D.~Held, C.~Mertz, and M.~Hebert.
\newblock Pcn: Point completion network.
\newblock {\em 2018 International Conference on 3D Vision (3DV)}, pages
  728--737, 2018.

\bibitem{Zhang2018EfficientSS}
J.~Zhang, H.~Zhao, A.~Yao, Y.~Chen, L.~Zhang, and H.~Liao.
\newblock Efficient semantic scene completion network with spatial group
  convolution.
\newblock In {\em European Conference on Computer Vision (ECCV)}, 2018.

\bibitem{Zhang2019CascadedCP}
P.~Zhang, W.~Liu, Y.~Lei, H.~Lu, and X.~Yang.
\newblock Cascaded context pyramid for full-resolution {3D} semantic scene
  completion.
\newblock {\em International Conference on Computer Vision (ICCV)}, pages
  7800--7809, 2019.

\bibitem{Zimmermann2017LearningFA}
K.~Zimmermann, T.~Petr{\'i}cek, V.~Salansk{\'y}, and T.~Svoboda.
\newblock Learning for active {3D} mapping.
\newblock {\em International Conference on Computer Vision (ICCV)}, pages
  1548--1556, 2017.

\end{thebibliography}
}

\definecolor{car}{rgb}{0.39215686, 0.58823529, 0.96078431}
\definecolor{bicycle}{rgb}{0.39215686, 0.90196078, 0.96078431}
\definecolor{motorcycle}{rgb}{0.11764706, 0.23529412, 0.58823529}
\definecolor{truck}{rgb}{0.31372549, 0.11764706, 0.70588235}
\definecolor{other-vehicle}{rgb}{0.39215686, 0.31372549, 0.98039216}
\definecolor{person}{rgb}{1.        , 0.11764706, 0.11764706}
\definecolor{bicyclist}{rgb}{1.        , 0.15686275, 0.78431373}
\definecolor{motorcyclist}{rgb}{0.58823529, 0.11764706, 0.35294118}
\definecolor{road}{rgb}{1.        , 0.        , 1.        }
\definecolor{parking}{rgb}{1.        , 0.58823529, 1.        }
\definecolor{sidewalk}{rgb}{0.29411765, 0.        , 0.29411765}
\definecolor{other-ground}{rgb}{0.68627451, 0.        , 0.29411765}
\definecolor{building}{rgb}{1.        , 0.78431373, 0.        }
\definecolor{fence}{rgb}{1.        , 0.47058824, 0.19607843}
\definecolor{vegetation}{rgb}{0.        , 0.68627451, 0.        }
\definecolor{trunk}{rgb}{0.52941176, 0.23529412, 0.        }
\definecolor{terrain}{rgb}{0.58823529, 0.94117647, 0.31372549}
\definecolor{pole}{rgb}{1.        , 0.94117647, 0.58823529}
\definecolor{traffic-sign}{rgb}{1.        , 0.        , 0.    }    

\section*{Supplementary}
\appendix

\begin{figure*}
	\centering
	\centering
	\scriptsize
	\setlength{\tabcolsep}{0.017\linewidth}
	\renewcommand{\arraystretch}{0.8}
	\begin{tabular}{cccc}

		\multicolumn{4}{c}{\textbf{\footnotesize Performance in SemanticKITTI \cite{Behley2019SemanticKITTIAD} (64 layers)}} \vspace{0.3cm}\\
		
		Input & SSCNet-full~\cite{Song2017SemanticSC} & LMSCNet (ours) & Ground Truth \\
		
		\includegraphics[width=0.4\columnwidth]{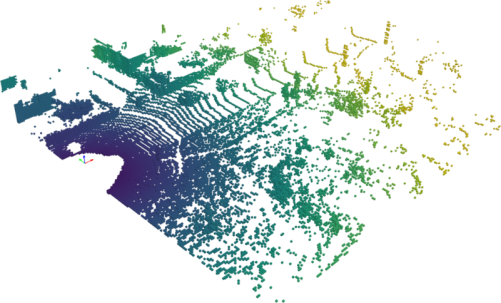} & 
		\includegraphics[width=0.4\columnwidth]{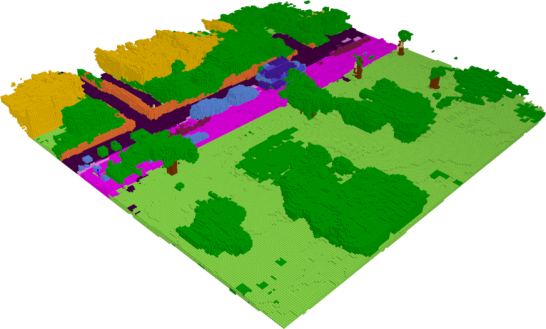} &
		\includegraphics[width=0.4\columnwidth]{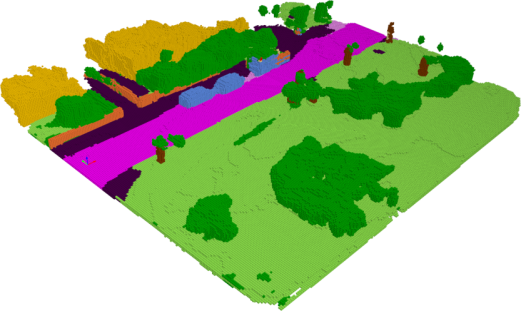} &
		\includegraphics[width=0.4\columnwidth]{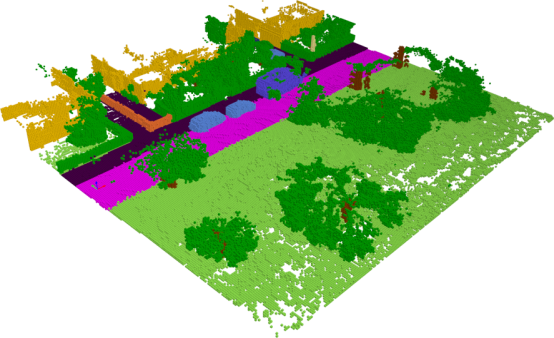} \\
		
		\includegraphics[width=0.4\columnwidth]{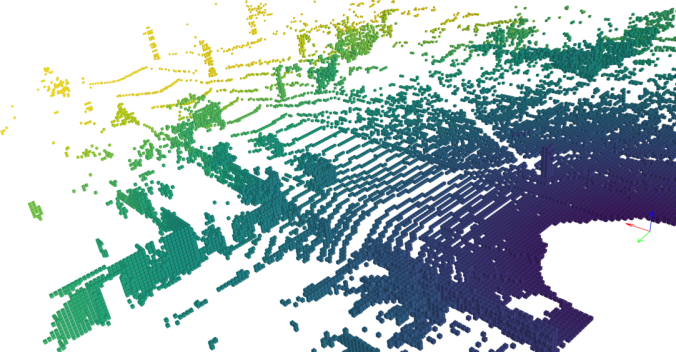} & 
		\includegraphics[width=0.4\columnwidth]{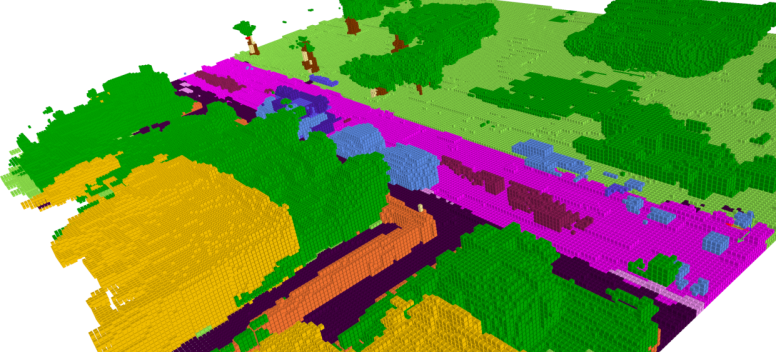} &
		\includegraphics[width=0.4\columnwidth]{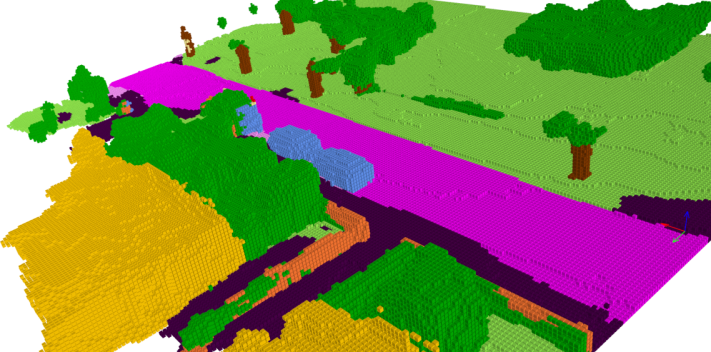} &
		\includegraphics[width=0.4\columnwidth]{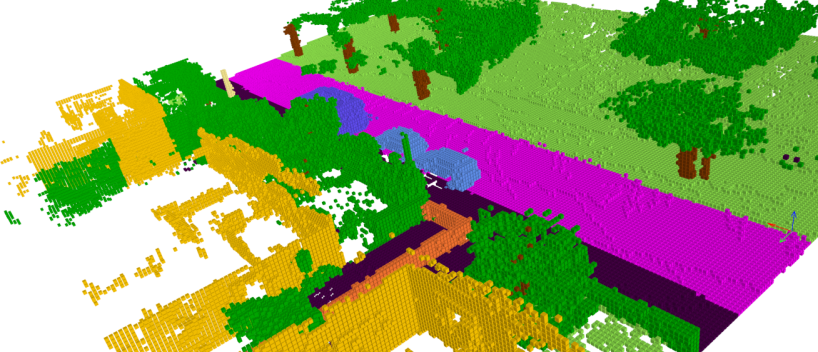} \vspace{0.5cm}\\

		\includegraphics[width=0.4\columnwidth]{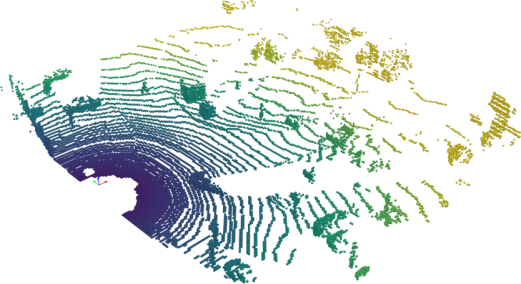} & 
		\includegraphics[width=0.4\columnwidth]{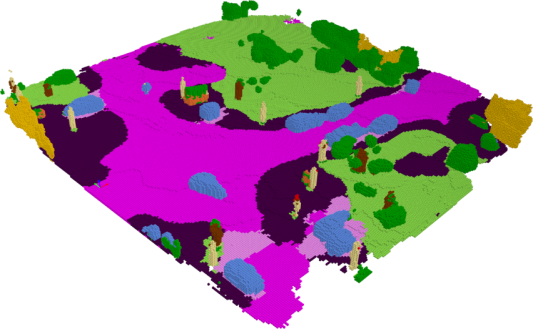} &
		\includegraphics[width=0.4\columnwidth]{Figures/method_qualitative/003985/003985_01_prours.png} &
		\includegraphics[width=0.4\columnwidth]{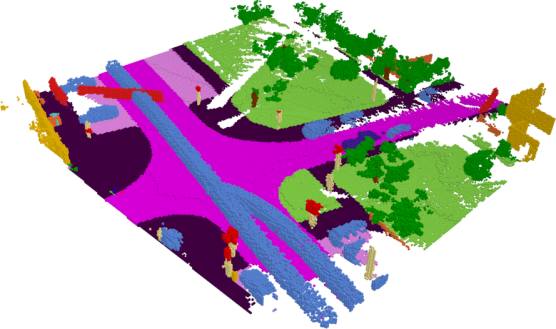} \\
		
		\includegraphics[width=0.4\columnwidth]{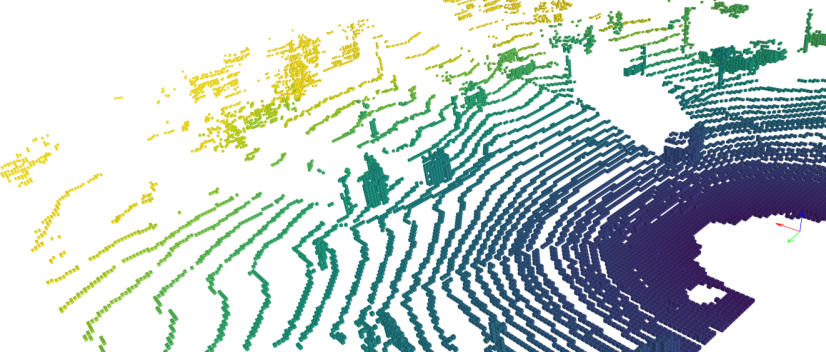} & 
		\includegraphics[width=0.4\columnwidth]{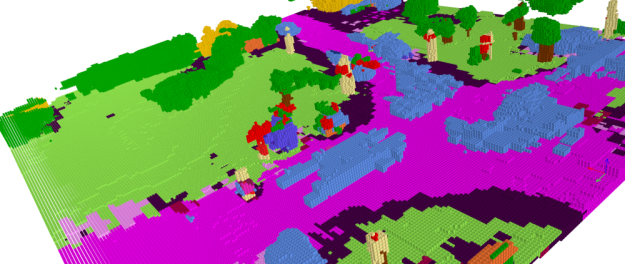} &
		\includegraphics[width=0.4\columnwidth]{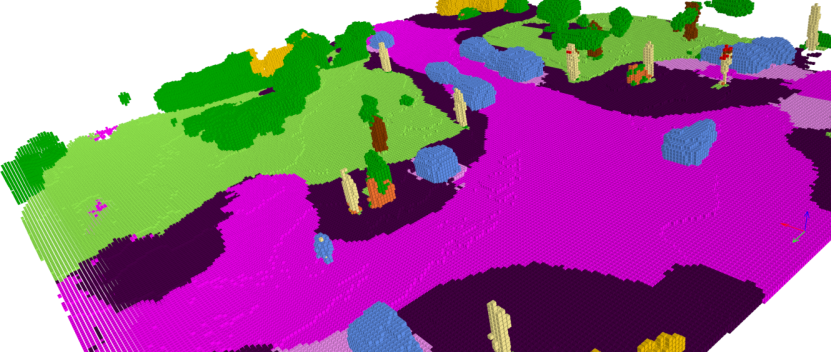} &
		\includegraphics[width=0.4\columnwidth]{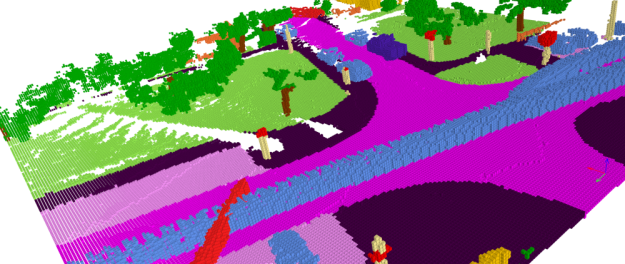} \vspace{0.5cm}\\ 
		
	\end{tabular}
	\label{fig:supp-qualitative-SemKITTI}
	
	\centering
	\scriptsize
	\setlength{\tabcolsep}{0.035\linewidth}
	\renewcommand{\arraystretch}{1.4}
	\begin{tabular}{ccc}

		\multicolumn{3}{c}{\textbf{\footnotesize Performance in nuScenes \cite{Caesar2019nuScenesAM} (32 layers)}} \vspace{0.15cm}\\
		
		Input & SSCNet-full~\cite{Song2017SemanticSC} & LMSCNet (ours)\\
		
		\includegraphics[width=0.4\columnwidth]{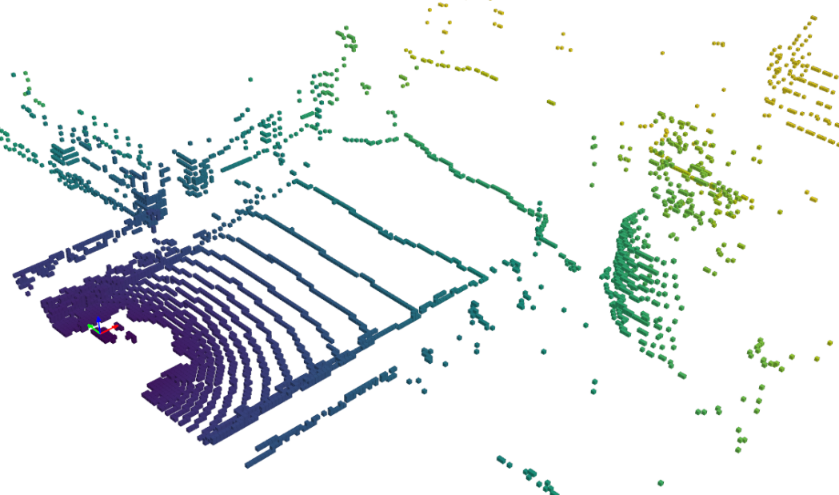} & 
		\includegraphics[width=0.4\columnwidth]{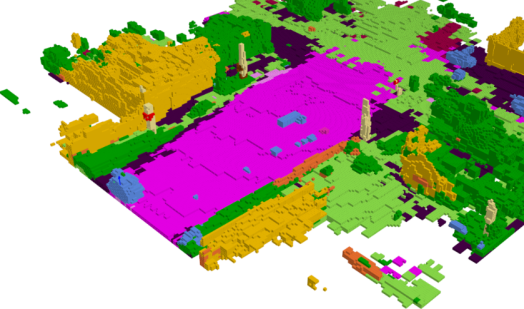} &
		\includegraphics[width=0.4\columnwidth]{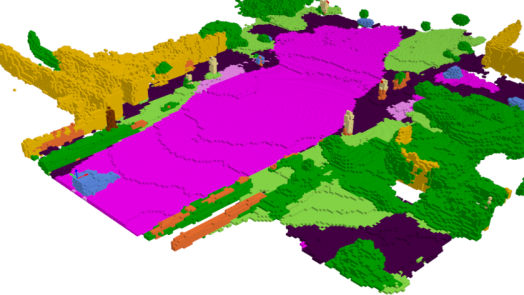}\\
		
		\includegraphics[width=0.4\columnwidth]{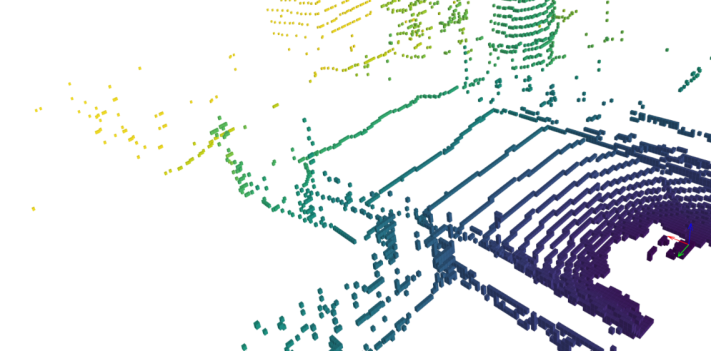} & 
		\includegraphics[width=0.4\columnwidth]{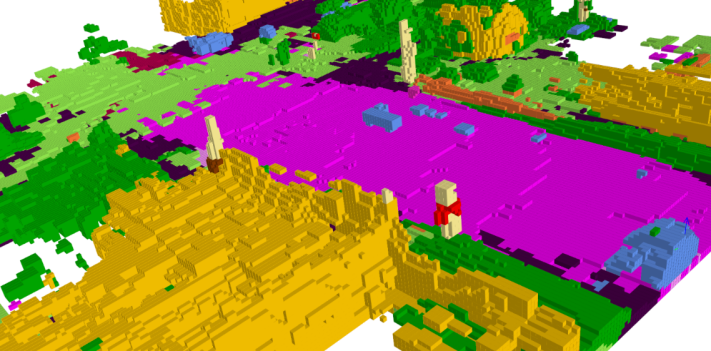} &
		\includegraphics[width=0.4\columnwidth]{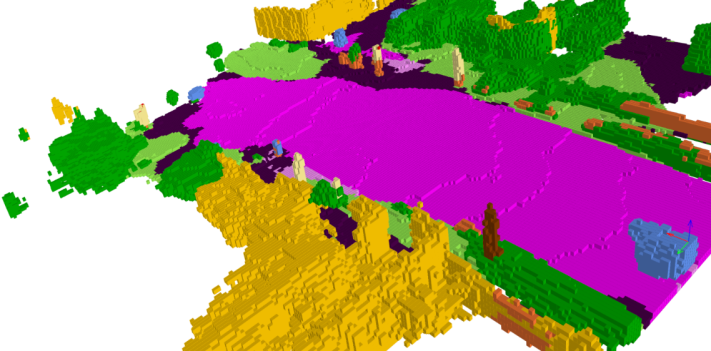}\\ \vspace{0.2cm}\\

		\includegraphics[width=0.4\columnwidth]{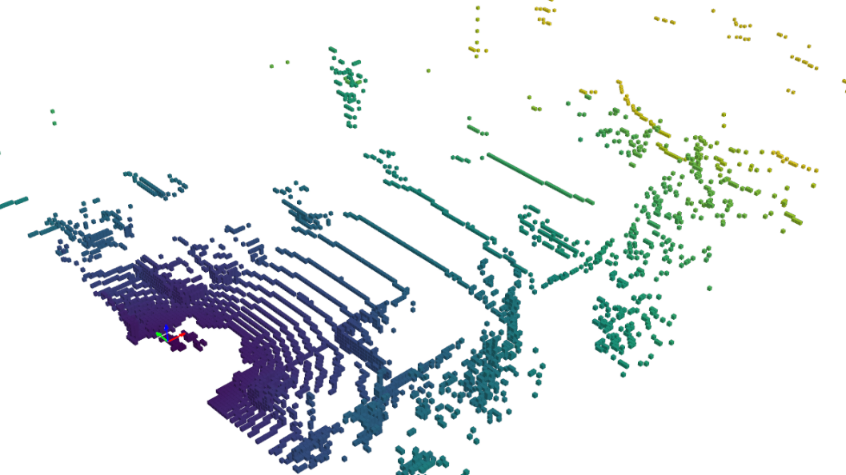} & 
		\includegraphics[width=0.4\columnwidth]{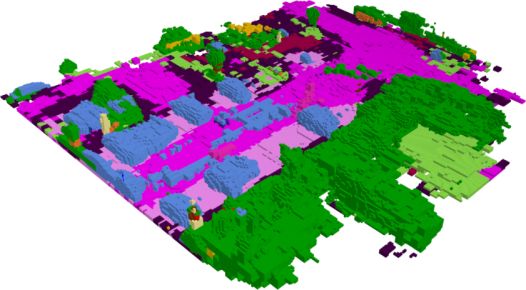} &
		\includegraphics[width=0.4\columnwidth]{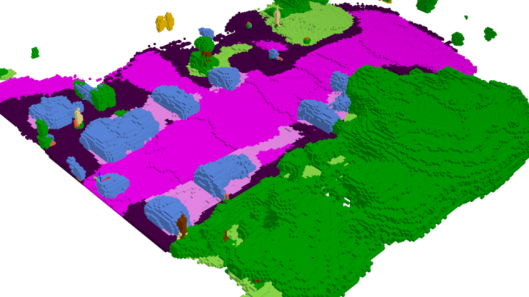}\\
		
		\includegraphics[width=0.4\columnwidth]{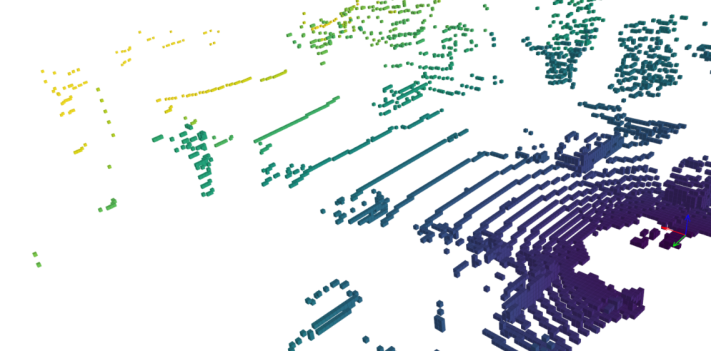} & 
		\includegraphics[width=0.4\columnwidth]{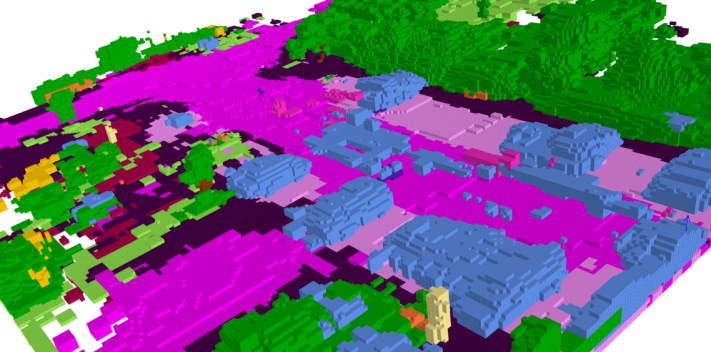} &
		\includegraphics[width=0.4\columnwidth]{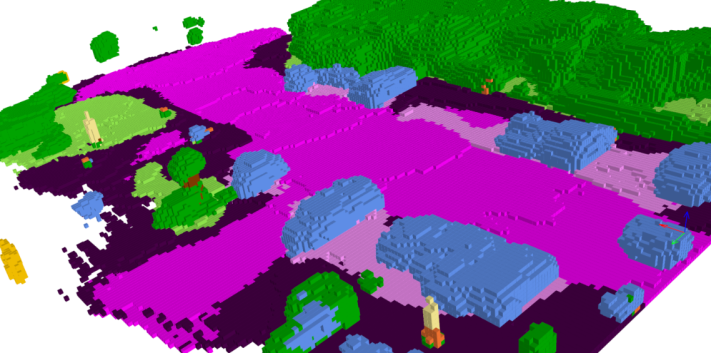}\\ \vspace{0.05cm}\\
		
	\end{tabular}
	\label{fig:supp-qualitative-nuScenes}
	
	\centering  
	\setlength{\tabcolsep}{0.017\linewidth}
	\renewcommand{\arraystretch}{0.8}
	\begin{tabular}{cccc}
		\multicolumn{4}{c}{\includegraphics[width=1.7\columnwidth]{Figures/method_qualitative/legend_method_qualitative.pdf}}
	\end{tabular}
	
	\caption{Additional qualitative 3D semantic completion results of our method in both SemanticKITTI \cite{Behley2019SemanticKITTIAD} (top rows) and nuScenes \cite{Caesar2019nuScenesAM} (bottom rows). Each pair of rows shows a single scene with different viewpoints. Groundtruth images are not shown for nuScenes due to the absence of point-wise semantic labels.}
	\label{fig:supp-qualitative1}
\end{figure*}

\section{Technical details}
\label{sec:supp-technical}

\subsection{Baselines implementations}
\label{sec:supp-technical-baselines}
Hereafter, we provide additional details on the implementation of the baselines listed in main article Tab. \textcolor{red}{1}.

\paragraph{TS3D baselines.} We compare our method with 3 variants of the Two Stream 3D (TS3D) network as reported in~\cite{Behley2019SemanticKITTIAD}, which originate from~\cite{Garbade2019TwoS3}. As in their original work, TS3D uses an additional RGB modality, TS3D+DNet and TS3D+DNet+RangeNet use instead LiDAR intensity. The network is modified in two ways: first, by directly using projected semantic labels to the input grid obtained by a LiDAR-based semantic segmentation network~\cite{Milioto2019RangeNetF} (TS3D+DNet); and secondly, by exchanging the 3D-CNN backbone by SATNet~\cite{Liu2018SeeAT} (TS3D+DNet+SATNet). The semantic labels obtained from the 2D branch on the 3 variants are one-hot encoded and lifted to the 3D grid resulting on a (N+1)$\times$H$\times$W$\times$D input tensor.

\paragraph{SSCNet baselines.} Following the practice in~\cite{Behley2019SemanticKITTIAD}, we use SSCNet~\cite{Song2017SemanticSC} without the flipped TSDF as input encoding. However, while \cite{Behley2019SemanticKITTIAD} only compares with SSCNet, which predictions are $1/4$ of the original input resolution, we also propose SSCNet-full, which outputs fullsize predictions. This is done by applying a 4$\times$4$\times$4 transpose convolution to the last layer of the network to retrieve original dimensions. For data balancing, we use their strategy by randomly subsampling occluded free space, conserving a 2:1 free-occupied ratio. 

\subsection{Architecture comparison}
\label{sec:supp-technical-architecture}

\begin{figure}
	\centering
	\includegraphics[width=0.76\linewidth]{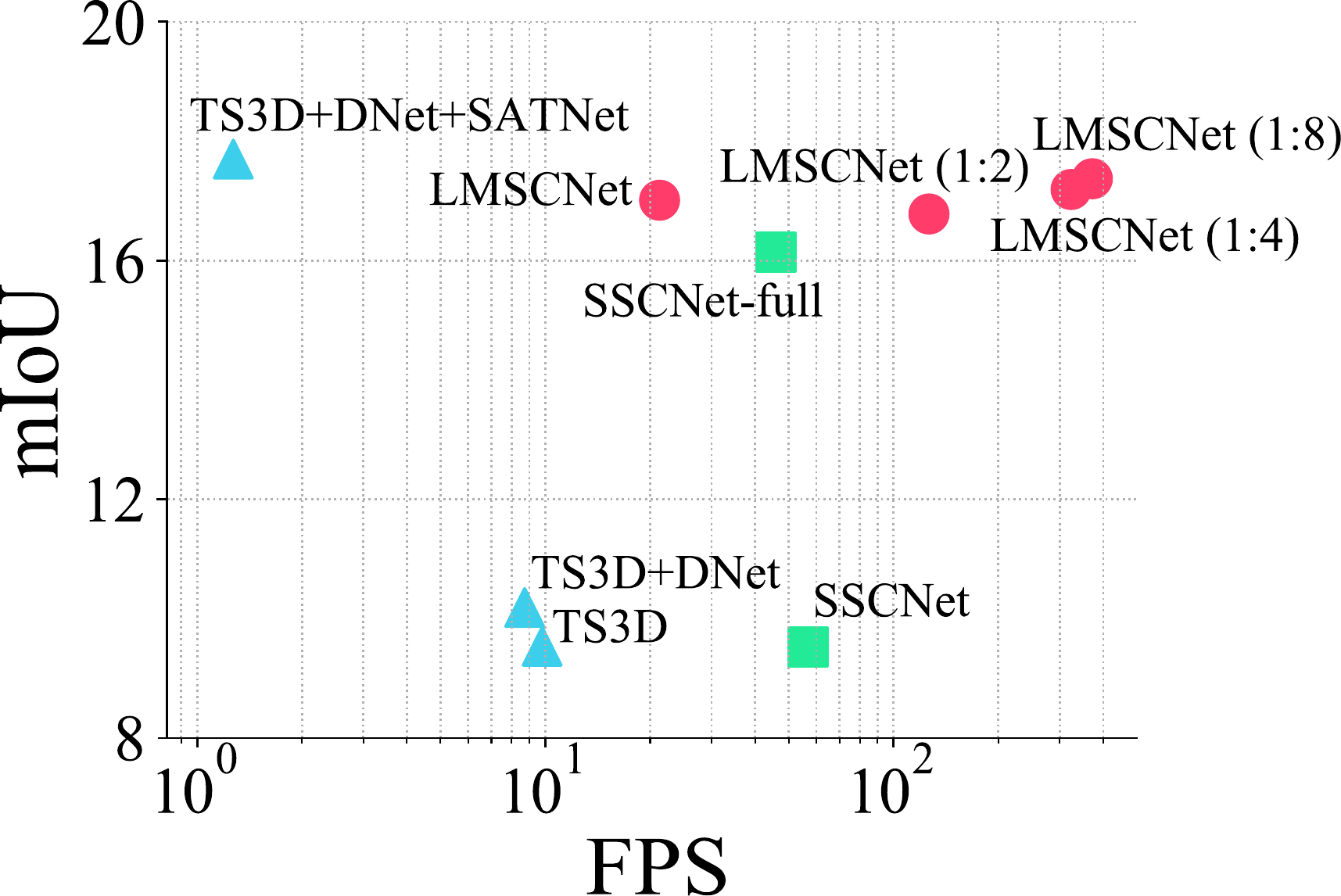}
	\caption{Network speed (FPS) and performance (mIoU). FPS are shown in log-scale. Notice that our method keeps good performance and fast inference time, this is specially noticeable for our coarse scale versions LMSCNet (1:x).}
	\label{fig:supp-tec-speedplot}
\end{figure}

Fig.~\ref{fig:supp-tec-speedplot} provides additional insight about the benefit of each architecture, where the top rightmost corresponds to the best speed-performance trade-off. Even though SSCNet-full achieves faster inference than our method at the original scale, the performance is slightly lower and more noisy as observed in Fig. \ref{fig:supp-qualitative1}. TS3D+DNet+RangeNet achieves slightly higher performance but the inference time and the number of parameters are considerably higher as seen in main article Tab. \textcolor{red}{3}. Considering this, our network keeps the best speed-performance trade-off. The interest of the coarser scale inferences of our method can be highlighted by the considerably lower inference times and high performances. 

\section{Qualitative results}
\label{sec:supp-qualitative}

In Fig. \ref{fig:supp-qualitative1} further qualitative results of our method in both SemanticKITTI \cite{Behley2019SemanticKITTIAD} and nuScenes \cite{Caesar2019nuScenesAM} datasets are provided. Notice our network performs smoother and less noisy reconstructions. Even though the ground-truth in SemanticKITTI accumulates scans of dynamic objects as seen in rows 3-4, our network reconstructs the vehicles correctly. This can be explained by the abundance of parked vehicles in the dataset. Performance in nuScenes can be seen in rows 5 to 8. It can be observed again the smoothness of the reconstruction when compared to SSCNet-full, with less noisy objects in the middle of the road. Notice that the network has been trained on SemanticKITTI, which explains the high vegetation predictions in nuScenes.
We refer the reader to the supplementary video for more qualitative insights.

\end{document}